%% file: main.tex
\documentclass[]{fairmeta}
\microtypesetup{expansion=false}

\usepackage{amsmath}
\usepackage{amssymb}
\usepackage{mathtools}
\usepackage{amsthm}
\usepackage{siunitx}
\usepackage{pifont}
\usepackage{adjustbox}
\usepackage{makecell}
\usepackage{wrapfig}
\usepackage{algorithm}
\usepackage{algorithmicx}
\usepackage{algpseudocode}
\usepackage{listings}
\usepackage{manyfoot}
\usepackage[title]{appendix}
\usepackage{mathrsfs}
\usepackage{colortbl}
\usepackage{array}
\usepackage{tabularx}


\definecolor{wfheader}{RGB}{213,229,241}
\definecolor{wftotalrow}{RGB}{245,235,217}
\definecolor{wfaveragerow}{RGB}{213,233,218}
\definecolor{wfapi}{RGB}{232,243,252}
\definecolor{wfapihead}{RGB}{211,231,245}
\definecolor{wfopen}{RGB}{235,247,234}
\definecolor{wfopenhead}{RGB}{216,237,216}
\definecolor{wfhuman}{RGB}{243,237,250}
\definecolor{wfgreen}{RGB}{220,239,222}
\definecolor{wfgreenhead}{RGB}{191,222,195}
\definecolor{wfdelta}{RGB}{239,247,241}
\definecolor{wfred}{RGB}{250,229,225}
\definecolor{wfbest}{RGB}{45,105,76}
\definecolor{gaincolor}{HTML}{6F9474}
\definecolor{losscolor}{HTML}{B95F56}
\definecolor{neutralgray}{HTML}{777777}
\newcolumntype{O}{>{\columncolor{wfgreen}}r}
\newcolumntype{P}{>{\columncolor{wfapi}}r}
\newcolumntype{Q}{>{\columncolor{wfopen}}r}
\newcolumntype{R}{>{\columncolor{wfgreen}}r}
\newcolumntype{T}{>{\columncolor{wfopen}}r}
\newcolumntype{U}{>{\columncolor{wfapi}}r}
\newcolumntype{K}{>{\columncolor{wfapihead}}r}
\newcolumntype{A}{>{\columncolor{wfdelta}}r}
\newcommand{\gain}[1]{\textcolor{gaincolor}{\bfseries #1}}
\newcommand{\loss}[1]{\textcolor{losscolor}{#1}}
\newcommand{\same}[1]{\textcolor{neutralgray}{#1}}
\newcommand{\best}[1]{\textcolor{wfbest}{\bfseries #1}}

\newcommand{\na}{\textemdash}

\theoremstyle{plain}

\theoremstyle{definition}

\theoremstyle{remark}

\title{EmbodiedSkills: World Action Model-augmented Cascaded Skills for Vision-Language-Action Agent}

\newcommand{\authorentry}[2]{%
  \author[#1]{#2}%
}

\authorentry{1}{Wei Wang}
\authorentry{1,\ddagger}{Wenqiao Zhang}
\authorentry{1}{Yutong Lin}
\authorentry{1}{Yuqian Yuan}
\authorentry{1}{Tianwei Lin}
\authorentry{1}{Jinhao Mao}
\authorentry{1}{Zhenxuan Fan}
\authorentry{1}{Mingjian Gao}
\authorentry{1}{Yang Dai}
\author[2]{Wentong Li}
\author[3]{Zheqi Lv}
\author[4]{Zheng Dong}
\author[6]{Yingjie Niu}
\author[5]{Jiaqi Zhu}
\authorentry{1}{Jun Xiao}
\author[6]{Chao Li}
\authorentry{1}{Yueting Zhuang}

\affiliation[1]{College of Computer Science and Technology, Zhejiang University}
\affiliation[2]{Nanjing University of Aeronautics and Astronautics}
\affiliation[3]{Cornell University}
\affiliation[4]{Universal Ubiquitous AI Co., Ltd.}
\affiliation[5]{National University of Singapore}
\affiliation[6]{Hangzhou DEEP Robotics Technology Co., Ltd.}
\contribution[\ddagger]{Corresponding author}

\abstract{
Vision--language--action (VLA) policies generate executable robot actions, but plausible proposals can have different consequences for the task. We present \textbf{EmbodiedSkills}, a skill-based agent framework that augments frozen VLA policies with world action models for consequence-aware action selection. The framework defines reusable skills for scene observation, context encoding, candidate generation, consequence prediction, outcome evaluation, candidate comparison, and action execution. Five stages organize their input--output dependencies in a shared feedback loop, accommodating both continuous action chunks and pick-and-place primitives. At each decision point, independently trained world action models predict the latent consequences of a common set of policy proposals. Selection combines their outcome scores and prediction disagreement with a lightweight comparator that estimates the benefit of an alternative from agreement in its predicted effects. The predictive and comparison modules learn from matched-state candidate branches, while the action policy and pretrained observation encoders remain frozen. On MetaWorld MT50 and CLIPort, EmbodiedSkills improves success rates from 76.28\% to 78.64\% and from 33.94\% to 36.83\%, respectively. An additional CALVIN evaluation with the frozen FLOWER policy increases five-task-chain success from 77.10\% to 77.90\%. These results demonstrate the applicability of consequence-aware selection across different policy backends and action representations.
}

\begin{document}

\maketitle

\input{sections/1.introduction}
\input{sections/2.related_work}
\input{sections/3.methodology}
\input{sections/4.agentic_rl}
\input{sections/5.experiments}
\input{sections/6.limitations}
\input{sections/7.conclusion}

\FloatBarrier
\bibliographystyle{unsrtnat}
\bibliography{main}

\clearpage
\appendix
\input{sections/8.detailed_results}

\input{sections/9.expanded_comparisons}

\end{document}

%% file: sections/1.introduction.tex
\section{Introduction}\label{sec1}

\begin{figure*}[t]
\centering
\includegraphics[width=\linewidth]{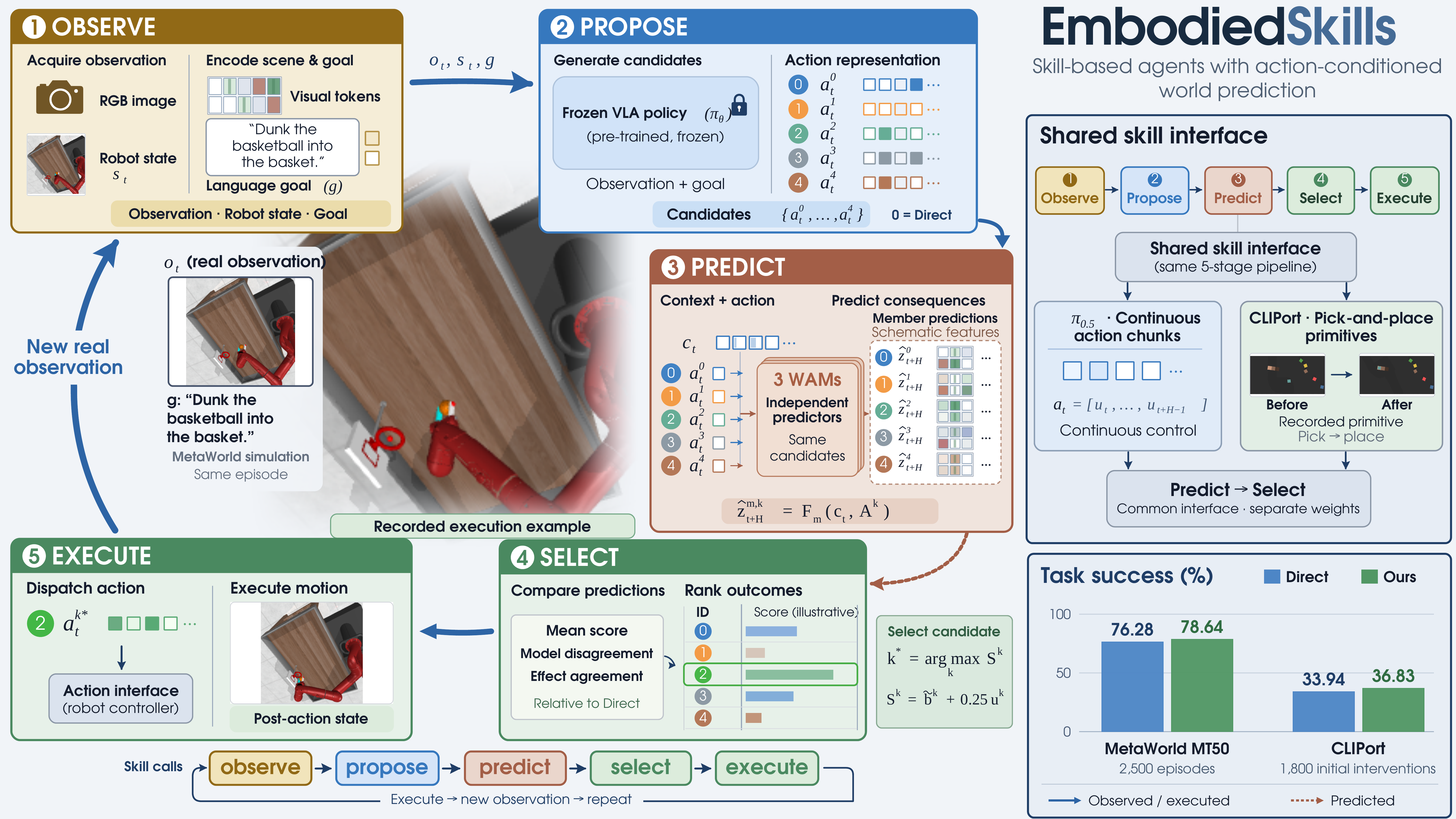}
\vspace{-3mm}
\caption{EmbodiedSkills augments a frozen action policy through five stages: \textsc{Observe}, \textsc{Propose}, \textsc{Predict}, \textsc{Select}, and \textsc{Execute}. Three independently trained world action models predict consequences for the same candidate actions. Selection combines their scores, disagreement, and predicted-effect agreement before dispatching the selected action. Shared skill interfaces accommodate continuous action chunks and manipulation primitives. Robot images show a recorded MetaWorld basketball episode; latent features and candidate scores are schematic. The result panel summarizes complete-task success on MetaWorld MT50 and CLIPort; their intervention protocols are detailed in Section~\ref{sec:eval-protocol}.}
\label{fig:intro}
\vspace{-2mm}
\end{figure*}

Vision--language--action (VLA) policies provide a reusable foundation for instruction-conditioned robot control, translating visual observations and language goals into executable actions~\cite{pmlr-v229-zitkovich23a,pmlr-v270-kim25c,pmlr-v305-black25a}. Generative policies can represent multiple ways to act from the same observation, including alternative motion trajectories and action chunks~\cite{chi2024diffusionpolicy,black2024pi0visionlanguageactionflowmodel}. This capability creates a decision problem at execution time: plausible actions can lead to different physical outcomes. When placing an object, for example, small differences in approach or release position can determine whether it reaches the intended location. Generating candidate actions alone does not establish which candidate is more likely to achieve the goal. This motivates a method that augments VLA policies with the ability to anticipate and compare candidate outcomes before execution.

Action-conditioned world models offer a way to anticipate these outcomes. Visual foresight predicts future observations under proposed actions~\cite{finn2017deep}, while latent world models such as DINO-WM and V-JEPA~2 support prediction and planning in learned representations~\cite{zhou2025dinowm,assran2025vjepa2}. Recent policy-steering methods, including GPC, FOREWARN, and DREAMSTEER, further demonstrate how predicted consequences can inform the selection of actions proposed by an existing policy~\cite{qi2026inference,wu2025foresight,cui2026dreamsteer}. These advances motivate combining the action-generation capability of VLA policies with the predictive capability of world models. The resulting decisions also depend on the reliability of the forecasts: a high score from one predictor may reflect an inaccurate consequence estimate. Making this combination reusable requires a framework that accommodates different policy backends and world models while preserving a consistent decision and execution workflow as components change. Skill-based robotic agents offer a useful foundation for organizing such capabilities~\cite{pmlr-v205-ichter23a}. Representing observation, action proposal, prediction, selection, and execution as callable skills with shared interfaces allows their implementations to be adapted or replaced within a common feedback loop.

Following this design, we introduce \textbf{EmbodiedSkills}, a skill-based agent framework that augments frozen VLA policies with world action models. An embodied skill is a reusable operation with defined inputs and outputs in the perception--decision--action loop. Five stages organize these operations: \textsc{Observe} uses scene observation and context encoding; \textsc{Propose} uses candidate generation; \textsc{Predict} uses consequence prediction and outcome evaluation; \textsc{Select} uses candidate comparison; and \textsc{Execute} uses action execution. Stages specify the information dependencies, while skills supply the capabilities that satisfy them. Their interfaces preserve the correspondence between a proposed action, its predicted effect, and the action sent to the environment. Policy-specific action representations and model-specific predictive features are handled within the corresponding skills, allowing implementations to be adapted or replaced while preserving the five-stage workflow.

Within this loop, three independently trained world action models predict latent consequences and goal-conditioned outcome scores for the same policy proposals. Selection first aggregates their scores while accounting for disagreement, then applies a lightweight learned correction based on agreement in the predicted effects. The comparator estimates whether replacing the policy's direct proposal is likely to improve or damage the outcome. Prediction and comparison are supervised by candidate branches executed from matched initial states: observed transitions provide consequence targets, and branch outcomes provide action-quality labels. During deployment, the models consume current observations and proposed actions; future observations and branch outcomes are reserved for supervision and evaluation.

This design concentrates adaptation in additional predictive and comparison modules while keeping the base policy and pretrained observation encoders frozen. Latent prediction avoids pixel-level future reconstruction in the decision loop, and the skill abstraction accommodates both continuous control and manipulation primitives. We instantiate the framework with $\pi_{0.5}$ on MetaWorld MT50~\cite{pmlr-v305-black25a,yu2020metaworld} and with the language-conditioned CLIPort policy~\cite{shridhar2022cliport}. The framework increases complete-task success from 76.28\% to 78.64\% on MT50 and from 33.94\% to 36.83\% on CLIPort. Ablations examine consequence supervision, joint predictive learning, language conditioning, and model disagreement, while alternative visual representations demonstrate component reuse within the same workflow. These evaluations cover repeated chunk selection on MT50 and initial-primitive selection on CLIPort. Additional comparisons examine inference-time policy augmentation methods, and a CALVIN five-task-chain study~\cite{mees2022calvinbenchmarklanguageconditionedpolicy} extends the framework to the frozen FLOWER backend~\cite{reuss2025flower}, improving complete-chain success from 77.10\% to 77.90\%.

Our contributions are threefold:
\begin{itemize}
    \item \textbf{A skill-based loop for predictive action selection.} Five stages compose observation, encoding, generation, prediction, evaluation, comparison, and execution skills through explicit input--output relationships, preserving candidate identity from policy proposal to physical execution.
    \item \textbf{Branch-supervised prediction and consequence comparison.} Independently trained world action models provide candidate outcomes and scores; a shared comparator uses predicted-effect agreement to refine disagreement-aware selection relative to the direct policy proposal.
    \item \textbf{Modular integration of policy backends and world models.} Shared skill interfaces support the adaptation and replacement of action-policy backends and world action models, accommodating different action representations and predictive features within the same agent workflow.
\end{itemize}

%% file: sections/2.related_work.tex
\section{Related Work}\label{sec2}

\subsection{Vision--Language--Action Policies}
Vision--language--action policies connect visual observations and language instructions to robot control. RT-2 transfers vision--language knowledge to robotic actions, while OpenVLA and $\pi_{0.5}$ advance reusable instruction-conditioned policies~\cite{pmlr-v229-zitkovich23a,pmlr-v270-kim25c,pmlr-v305-black25a}. Diffusion Policy and $\pi_0$ use diffusion and flow matching to generate continuous action sequences~\cite{chi2024diffusionpolicy,black2024pi0visionlanguageactionflowmodel}, while CLIPort combines semantic and spatial pathways to predict pick-and-place actions~\cite{shridhar2022cliport}. These policies provide executable proposals, but generating plausible actions does not establish which candidate will best achieve the goal. We augment frozen policies with consequence prediction and candidate selection to inform this decision.

\subsection{Action-Conditioned World Models}
World models learn how an environment changes under an action, providing a basis for evaluating future behavior. Deep Visual Foresight uses action-conditioned video prediction for robotic planning~\cite{finn2017deep}. IRASim strengthens the alignment between actions and video frames to model detailed robot--object interactions and support model-based planning~\cite{zhu2025irasim}. Latent approaches predict changes in learned visual representations. DINO-WM learns action-conditioned transitions over pretrained visual features, and V-JEPA~2 combines self-supervised video learning with action-conditioned post-training for robotic planning~\cite{zhou2025dinowm,assran2025vjepa2}. LaDi-WM predicts future geometric and semantic features and incorporates these forecasts into action refinement~\cite{huang2025ladiwm}.

Accurate prediction alone does not determine whether a future state serves the task goal, and errors in the predicted consequences can affect subsequent selection. EmbodiedSkills connects latent prediction with goal-conditioned evaluation and uses disagreement and effect agreement across independently trained predictors when comparing candidates.

\subsection{Predictive Action Selection}
Improving an existing policy through inference-time selection has been explored with both direct evaluators and predictive models. V-GPS reranks actions using a value function learned through offline reinforcement learning and supports multiple policy architectures without updating their weights~\cite{nakamoto2025vgps}. RoboMonkey combines test-time action sampling with a learned vision--language verifier~\cite{kwok2025robomonkey}. These approaches establish candidate evaluation as a practical mechanism for augmenting robot policies. KeyStone selects an existing proposal through geometric self-consistency without a learned evaluator~\cite{dai2026geometryguidedselfconsistencyphysical}. UF-OPS learns a contrastive verifier for on-policy steering, while GAF uses a learned action-value gradient during flow sampling~\cite{attarian2026updatefreeonpolicysteeringverifiers,yang2026guidedactionflowvalueguided}. SPIRAL combines reward modeling and residual policy improvement in the SARM2 framework~\cite{chen2026sarm2multitaskstageaware}; its critic also provides a component-level alternative to consequence prediction.

Recent lightweight evaluators further exploit signals available from a frozen policy. TACO estimates support for action chunks with a coin-flipping pseudo-count network and selects samples associated with successful behavior~\cite{yang2025taco}. Linear value probes can also extract success-related information from frozen VLA representations and use it to rank sampled action prefixes~\cite{zhang2026frozenvalue}. FabriMAE uses Markov Attention Entropy (MAE) to score sampled $\pi_{0.5}$ actions from the policy's own visual attention~\cite{aniri2026fabrimae}. Such readouts offer direct alternatives to learning action-conditioned future representations.

VGAS trains a chunk-level critic with geometric regularization for best-of-$N$ selection under few-shot adaptation~\cite{xu2026vgas}. Q-Planning learns an off-policy action-chunk value function beside a frozen FastWAM policy and updates that value function from deployment rollouts~\cite{giridhar2026qplanning}. VLA-Corrector acts at a different point in the control loop: a learned latent dynamics monitor interrupts an action chunk when its observed execution diverges from prediction, then guides replanning~\cite{pan2026vlacorrector}. These methods reinforce the value of inference-time policy augmentation while using different training data, candidate generators, or intervention timing. Execution-focused methods address complementary concerns: RTC and VLA-RAIL coordinate chunk execution, AutoHorizon and A$^3$ adapt execution commitment, and ASRR refines proposed action sequences~\cite{black2025realtimeexecutionactionchunking,zhao2025vlarailrealtimeasynchronousinference,wang2026vlaknowslimitsadaptive,chen2026dynamicexecutioncommitmentvisionlanguageaction,asrr2026}. Our comparisons adapt author implementations to the local backends and distinguish complete methods from reused components and changes in execution schedule.

Predictive approaches additionally model the consequences of candidate actions. GPC couples a frozen diffusion policy with an action-conditioned world model to rank or refine proposals~\cite{qi2026inference}. Latent Policy Steering combines a world model pretrained across embodiments with a learned value function to select among behavior-cloned policy candidates~\cite{wang2025latentpolicy}. FOREWARN aligns a vision--language model with predicted latent futures to assess and filter low-level plans~\cite{wu2025foresight}, while DREAMSTEER evaluates imagined outcomes of action chunks from a frozen VLA and motion primitives using a language-conditioned value model~\cite{cui2026dreamsteer}. WEAVER learns multi-view latent dynamics and rewards with flow matching, enabling policy evaluation, improvement, and test-time planning with $\pi_{0.5}$~\cite{jain2026weaver}. However, extending predictive selection across heterogeneous policies and world models requires shared interfaces for proposals, predictions, and execution. EmbodiedSkills provides this connection through a common skill loop. Within this loop, selection combines ensemble outcome scores with a learned correction derived from agreement in predicted effects, using the direct policy proposal as a fixed reference.

\subsection{Skills and Robotic Agents}
Temporal abstraction provides a foundation for composing robot behavior, with options formalizing temporally extended actions in reinforcement learning~\cite{sutton1999between}. Language-driven robotic agents connect such capabilities to task-level decisions. SayCan combines language-model scores with learned affordances to select skills, and Inner Monologue incorporates environment feedback into planning~\cite{pmlr-v205-ichter23a,huang2022innermonologueembodiedreasoning}. Code as Policies composes perception and control functions into executable programs, while VoxPoser constructs language-conditioned value maps for manipulation~\cite{liang2023codepolicieslanguagemodel,huang2023voxposercomposable3dvalue}.

These systems primarily organize task-level reasoning and behavior composition; coordinating candidate generation, world-model prediction, and action selection requires interfaces within the decision process itself. We organize these decision capabilities as reusable skills with defined inputs and outputs. Five stages specify their dependencies, while policy and world-model implementations supply the corresponding capabilities within a shared feedback loop.

%% file: sections/3.methodology.tex
\section{Methodology}\label{sec3}

\begin{figure*}[t]
\centering
\includegraphics[width=\linewidth]{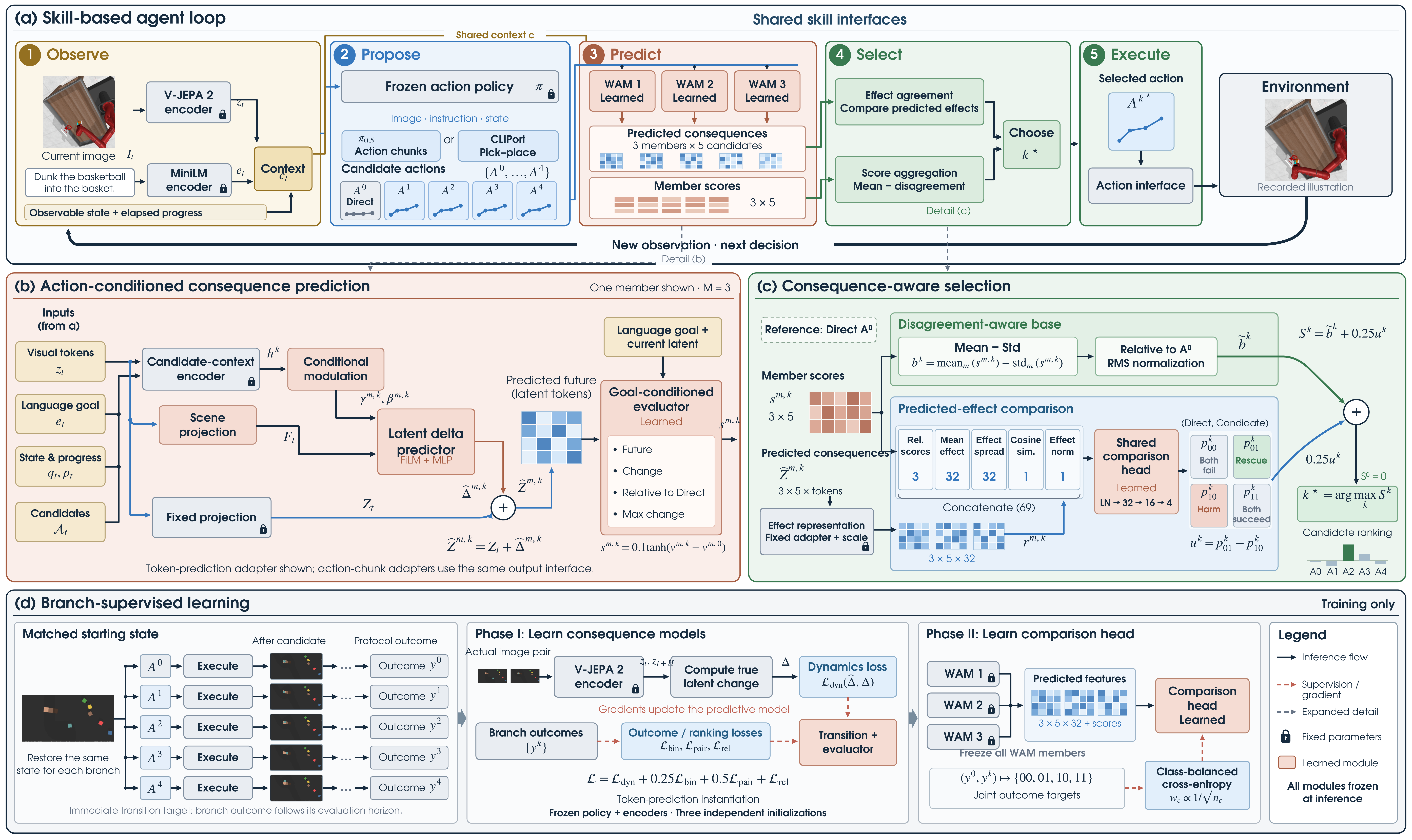}
\caption{EmbodiedSkills architecture. (a) Five stages organize observation, encoding, candidate-generation, prediction, evaluation, comparison, and execution skills in a shared feedback loop. Table~\ref{tab:stage-skills} defines their roles and outputs. (b) Each world action model predicts candidate-conditioned latent changes and evaluates the resulting consequences. (c) Selection combines ensemble scores with a learned comparison of predicted effects relative to the direct proposal. (d) Matched-state branches supervise the world action models, which are then frozen to train the comparison head. Future observations and branch outcomes provide training supervision only. Images show recorded MetaWorld observations and CLIPort candidate branches; the small ranking plot is schematic.}
\label{fig:wam-architecture}
\end{figure*}

EmbodiedSkills augments a frozen action policy with latent consequence prediction and candidate comparison. A five-stage skill loop separates the production of executable actions from the prediction and comparison of their effects. Figure~\ref{fig:wam-architecture} summarizes the framework. We first define this loop, then describe the world action models, the consensus-based selection rule, and branch-supervised learning.

\subsection{Skill-Based Decision Loop}\label{sec:skill-loop}

We define an embodied skill as a reusable operation with specified inputs and outputs in the robot's decision process. A \emph{stage} specifies when an operation is needed and which information must be available; its constituent \emph{skills} perform that operation. Table~\ref{tab:stage-skills} specifies the skills used in the five stages. These include acquiring observations, generating actions, and evaluating predicted effects, so that perception, reasoning about consequences, and physical execution can participate in the same composition. A skill can wrap an environment operation, a frozen model, or a learned module. Multiple skills may share a model forward pass, as in consequence prediction and outcome evaluation.

\input{tables/stage_skills}

At decision time $t$, the available context contains an image $I_t$, an instruction $g_t$, a sensor-derived state summary $q_t$, and elapsed execution progress $p_t$. Progress records the consumed execution budget. Frozen visual and language encoders produce $z_t=E_v(I_t)$ and $e_t=E_l(g_t)$, giving context $c_t=(z_t,e_t,q_t,p_t)$. The action policy $\pi$ supplies $K$ executable proposals,
\begin{equation}
    \mathcal A_t=\{A_t^0,\ldots,A_t^{K-1}\},\qquad K=5,
\end{equation}
where $A_t^0$ is the direct policy proposal. A candidate can represent a continuous action chunk or a manipulation primitive. Stochastic policies supply alternative samples, while spatial action policies expose distinct hypotheses from their native action distribution.

In \textsc{Observe}, scene observation acquires the available image and robot state, while context encoding constructs the visual and language representations. In \textsc{Propose}, candidate generation invokes the frozen policy and assigns persistent identities to its five proposals. In \textsc{Predict}, consequence prediction and outcome evaluation are instantiated by three independently trained WAMs. Each member assesses every proposal; together, they produce three forecasts and three outcome estimates per candidate. The five candidates come from the policy; the three members provide alternative estimates of their effects. In \textsc{Select}, candidate comparison combines these estimates with predicted-effect agreement and returns one candidate index. In \textsc{Execute}, action execution dispatches that candidate in the backend's native action representation. The stage-level compositions are
\begin{equation}
\begin{aligned}
 c_t&=\operatorname{Observe}(I_t,g_t,q_t,p_t),
 &\mathcal A_t&=\operatorname{Propose}(I_t,g_t,q_t),\\
 \mathcal P_t&=\operatorname{Predict}(c_t,\mathcal A_t),
 &k_t^\star&=\operatorname{Select}(\mathcal P_t),\\
 I_{t+H},q_{t+H}&=\operatorname{Execute}(A_t^{k_t^\star}).
\end{aligned}
\label{eq:skill-composition}
\end{equation}
Here $H$ is the executed horizon, which can denote a bounded action chunk or one completed primitive. Fresh feedback supplies the next iteration. These dependencies determine the stage order directly. Candidate identities are preserved across prediction, comparison, and execution, so the selected action remains exactly one of the policy proposals. Backend-specific action encoders and execution interfaces accommodate different action spaces without changing this loop or its selection rule.

The skill boundary fixes the meaning of an operation's output while allowing its implementation to change. For example, candidate generation can provide $\pi_{0.5}$ or FLOWER action chunks or CLIPort manipulation primitives; consequence prediction can use pooled or spatial visual latents. Each implementation must retain the correspondence between a candidate and its predicted effect, and candidate comparison must return an index into the original proposal set. This contract makes component substitution compatible with the downstream decision and execution stages. The experiments instantiate this composition across three policy backends and several predictive representations.

\subsection{Ensemble Consequence Prediction}\label{sec:latent-prediction}

We use a frozen V-JEPA~2 visual encoder~\cite{assran2025vjepa2} and a frozen MiniLM language encoder. Three independently initialized world action models share the same training data and observation representations. For member $m\in\{1,2,3\}$ and candidate $k$, prediction produces
\begin{equation}
 \widehat z_t^{m,k}=F_{\theta_m}(c_t,A_t^k;\mathcal A_t),
 \qquad
 s_t^{m,k}=V_{\psi_m}(\widehat z_t^{m,k},c_t).
\label{eq:ensemble-prediction}
\end{equation}
The first term forecasts the visual representation after executing the candidate; the second is an outcome score computed through that forecast. The candidate set is available to the context encoder so that it can represent an action relative to the other proposals. Scores are expressed on a consistent scale across members before aggregation. Language and observable state can condition consequence prediction as well as outcome evaluation.

The prediction skill separates pretrained visual encoding from the learned transition. In the spatial-token instantiation, a frozen candidate-context encoder conditions a lightweight token predictor. Let $h_t^k$ denote the resulting context and $Z_t=z_tP$ a fixed projection of current visual tokens. Candidate-specific modulation and latent change prediction take the form
\begin{equation}
\begin{aligned}
 (\gamma_t^{m,k},\beta_t^{m,k})&=W_m h_t^k,\\
 \widehat\Delta_t^{m,k}
 &=\mu_\Delta+\sigma_\Delta\odot D_m\!\left(
 B_m(z_t)\odot[1+0.1\tanh(\gamma_t^{m,k})]+\beta_t^{m,k}
 \right),\\
 \widehat Z_t^{m,k}&=Z_t+\widehat\Delta_t^{m,k}.
\end{aligned}
\label{eq:conditional-delta}
\end{equation}
The modulation is broadcast over spatial tokens. $B_m$ and $D_m$ are small feed-forward networks, and $\mu_\Delta,\sigma_\Delta$ are training-set statistics of observed latent changes. The evaluator combines summaries of the predicted future, its change from the current observation, its difference from the direct proposal's prediction, and the language goal. Its relative score is bounded as $s_t^{m,k}=0.1\tanh(v_t^{m,k}-v_t^{m,0})$, where $v_t^{m,k}$ is the evaluator output. Scores from the frozen candidate-context encoder are not added to this decision score.

Other action representations can use a different transition adapter while retaining the interface in Eq.~\eqref{eq:ensemble-prediction}. Continuous action chunks, for example, can be encoded recurrently and used to predict a pooled visual representation. The comparison stage consumes the same candidate-indexed scores and compact predicted effects. Observation encoding is shared across the three members, and predicting a candidate requires neither rendering a future image nor executing it in a simulator at deployment.

\subsection{Consequence Agreement for Action Selection}\label{sec:consequence-selection}

Selection combines a disagreement-aware score with a learned comparison against the direct proposal. We suppress the time index for clarity and use population standard deviations throughout. The base score is
\begin{equation}
 b^k=\frac{1}{M}\sum_{m=1}^{M}s^{m,k}
 -\sqrt{\frac{1}{M}\sum_{m=1}^{M}(s^{m,k}-\bar s^k)^2},
 \qquad M=3,
\label{eq:base-score}
\end{equation}
where $\bar s^k$ is the mean score. This discounts a candidate when its outcome estimates disagree. Relative scores are normalized within each candidate set:
\begin{equation}
 d^k=b^k-b^0,\qquad
 \widetilde b^k=\frac{d^k}
 {\max\!\left(\sqrt{\frac{1}{K-1}\sum_{j=1}^{K-1}(d^j)^2},\epsilon\right)}.
\label{eq:base-normalization}
\end{equation}

The comparison head also receives the predicted visual effects. A fixed representation adapter maps each member's predicted consequence relative to candidate $0$ to a compact effect $r^{m,k}\in\mathbb R^{32}$. For token predictions, spatial pooling gives a difference $\delta^{m,k}$ in the projected latent space, and $r^{m,k}=\delta^{m,k}\oslash\rho$, where $\rho$ is a per-dimension root-mean-square scale estimated from training alternatives. For higher-dimensional pooled predictions, the adapter normalizes the difference first and then applies training-fitted centering and a 32-dimensional principal-component projection. With $d=32$, we compute its member-wise mean $\bar r^k$, standard deviation $\sigma_r^k$, average pairwise directional agreement $a^k$, and average magnitude $\ell^k$:
\begin{equation}
 a^k=\frac{1}{3}\sum_{m<n}\langle\widehat r^{m,k},\widehat r^{n,k}\rangle,
 \qquad
 \ell^k=\frac{1}{M\sqrt d}\sum_m\|r^{m,k}\|_2,
\label{eq:effect-agreement}
\end{equation}
where $\widehat r$ denotes unit normalization with a numerical floor. The score features are $\widetilde s^{m,k}=(s^{m,k}-s^{m,0})/\eta$, with $\eta$ the root-mean-square relative score across all members and non-direct candidates. Both score normalization denominators use $\epsilon=10^{-6}$, and the fitted effect scale has a floor of $10^{-5}$.

These quantities form a 69-dimensional candidate descriptor,
\begin{equation}
 x^k=[\widetilde s^{1,k},\widetilde s^{2,k},\widetilde s^{3,k},
       \bar r^k,\sigma_r^k,a^k,\ell^k].
\end{equation}
A shared multilayer perceptron with layer normalization, hidden widths 32 and 16, GELU activations, and dropout 0.1 predicts four probabilities $p_{00}^k,p_{01}^k,p_{10}^k,p_{11}^k$. The subscripts denote the binary outcomes of the direct proposal and candidate $k$, respectively. Thus $p_{01}^k$ estimates a beneficial replacement and $p_{10}^k$ estimates a harmful one. The correction and final choice are
\begin{equation}
 u^k=p_{01}^k-p_{10}^k,\qquad
 S^k=\widetilde b^k+0.25u^k,\qquad
 k^\star=\arg\max_{k\in\{0,\ldots,K-1\}}S^k.
\label{eq:consensus-selection}
\end{equation}
We set $u^0=S^0=0$ and resolve ties by candidate order. The coefficient is fixed before evaluation. Candidate $0$ remains the fixed reference. The head uses only model-produced scores and predicted effects. Instruction and action information reach it through the upstream predictors.

\subsection{Branch-Supervised Learning}\label{sec:branch-learning}

Training examples compare alternative actions from the same simulator state. At each anchor, we save the current observation, instruction, observable state, and candidate set. Executing each candidate from the restored anchor supplies the observation after its execution horizon and an outcome label $y^k$ under a declared evaluation horizon. Prediction targets describe the immediate candidate transition. Outcome labels can instead describe a longer continuation when such branches are collected. Each backend's protocol specifies both horizons.

Each world action model learns from observed latent changes and action-quality supervision, keeping the base policy and pretrained encoders frozen. For the token predictor, normalized latent changes are supervised with a Smooth-$L_1$ loss. The evaluator combines a binary outcome loss, an ordering loss over candidates with different outcomes, and a Huber loss on the relative outcome $y^k-y^0$:
\begin{equation}
 \mathcal L_{\mathrm{WAM}}=
 \mathcal L_{\mathrm{dyn}}+0.25\mathcal L_{\mathrm{bin}}
 +0.5\mathcal L_{\mathrm{pair}}+\mathcal L_{\mathrm{rel}}.
\label{eq:wam-training}
\end{equation}
A dynamics-only warm-up precedes joint prediction and evaluation training. During joint training, evaluation gradients pass through the predicted latent consequence into the transition model. The three members use independent initializations, and their checkpoints are selected on internal validation outcomes. Backend adapters may use additional branch-quality targets while exposing the same prediction and scoring interface.

After this stage, all three members are frozen and the comparison head is trained separately. For each non-direct candidate, the target is the joint outcome class $c^k=2y^0+y^k$. Class weights are proportional to the inverse square root of training class counts and normalized to mean one. With $\mathcal B$ denoting the minibatch of anchor--candidate pairs, the weighted loss is
\begin{equation}
 \mathcal L_{\mathrm{cmp}}=
 -\frac{\sum_{(i,k)\in\mathcal B}w_{c_i^k}\log p_{c_i^k}(x_i^k)}
 {\sum_{(i,k)\in\mathcal B}w_{c_i^k}},\qquad k>0.
\label{eq:comparison-training}
\end{equation}
Task-balanced sampling limits domination by tasks with more collected anchors. The head checkpoint minimizes weighted validation cross-entropy, with earlier epochs resolving ties. Feature scales and any fitted projection are estimated from the training partition. At deployment, all learned components are frozen and decisions use the current context and policy proposals.

%% file: tables/stage_skills.tex
\begin{table}[!htb]
\centering
\small
\setlength{\tabcolsep}{5pt}
\renewcommand{\arraystretch}{1.18}
\begin{tabularx}{\linewidth}{@{}>{\raggedright\arraybackslash}p{.14\linewidth}>{\raggedright\arraybackslash}p{.25\linewidth}>{\raggedright\arraybackslash}X@{}}
\toprule
\rowcolor{wfheader}
\textbf{Stage} & \textbf{Constituent skills} & \textbf{Role and output} \\
\midrule
\textsc{Observe} & Scene observation;\newline context encoding & Acquire the current image and observable robot state; encode the image and instruction into the context used by prediction. \\
\textsc{Propose} & Candidate generation & Query the frozen action policy for five executable proposals, preserving candidate~0 as the direct-policy reference. \\
\textsc{Predict} & Consequence prediction;\newline outcome evaluation & Three WAM instances predict a latent consequence and an outcome score for each of the five candidates, using current context and proposed actions. \\
\textsc{Select} & Candidate comparison & Aggregate scores and their disagreement, compare predicted effects against candidate~0, and return the index of the highest-scoring proposal. \\
\textsc{Execute} & Action execution & Execute the indexed action chunk or manipulation primitive. The resulting observation supplies the next iteration. \\
\bottomrule
\end{tabularx}
\caption{Stage-specific skills in EmbodiedSkills. Stages define information flow; skills perform the constituent operations. The three WAMs instantiate the same prediction and evaluation skills with independent weights. Candidate identities are preserved through prediction, selection, and execution.}
\label{tab:stage-skills}
\end{table}

%% file: sections/4.agentic_rl.tex
\section{Policy and World Model Instantiations}\label{sec:agent_training}

The skills in Table~\ref{tab:stage-skills} support two distinct forms of action generation. On MetaWorld, the candidate-generation skill wraps a frozen $\pi_{0.5}$ policy~\cite{pmlr-v305-black25a} that generates continuous robot-control sequences. On CLIPort, it wraps the language-conditioned spatial policy~\cite{shridhar2022cliport}, whose proposals specify pick-and-place primitives. Both instantiations expose five candidates, preserve the direct proposal as candidate~0, and use the consequence-consensus rule in Eq.~\eqref{eq:consensus-selection}. Their action encoders and execution horizons differ, while observation, prediction, comparison, and candidate dispatch retain the same roles. Context encoding, consequence prediction, and action execution use adapters for their respective representations while preserving the stage dependencies.

\paragraph{Continuous control.}
We use the public MetaWorld-adapted $\pi_{0.5}$ checkpoint with RGB observations from the corner camera. The policy predicts ten actions per proposal, of which the first eight are scored and executed. Each action has four control dimensions. Five stochastic samples provide alternative chunks, with the sampling of candidate~0 aligned to the direct-policy baseline. The WAM conditions on pooled V-JEPA~2 features, MiniLM instruction features, observable robot state, and elapsed execution progress. A recurrent action encoder accommodates the continuous chunk and predicts its subsequent visual representation. The three members use hidden width~192. Candidate-relative predicted effects are reduced to 32 dimensions using a projection fitted on training data before entering the shared comparison head.

\paragraph{Spatial manipulation.}
The CLIPort candidate-generation skill obtains five distinct pick-and-place hypotheses from the frozen policy. Each candidate is executed as one complete manipulation primitive. The predictor operates on 64 spatial visual tokens, with a fixed random orthogonal projection from 1,024 to 32 dimensions. A frozen context encoder conditions the candidate-specific transition, whose hidden width is~128. The outcome evaluator derives its score from the predicted consequence, as defined in Section~\ref{sec:latent-prediction}.

\paragraph{Learning and component reuse.}
The CLIPort branch archive contains 1,290 anchors across the 18 task settings, partitioned by task and seed into 1,032 training anchors and 258 anchors for checkpoint selection. Prediction uses 20 warm-up epochs followed by 60 epochs of joint consequence and outcome learning. The joint-stage predictor and value-head learning rates are both $2\times10^{-4}$; the consequence-loss coefficient is one. MetaWorld uses 12,519 training anchors from 750 episodes and 4,284 validation anchors from 250 episodes. Its WAMs are trained for 30 epochs with batch size~128 and AdamW learning rate $3\times10^{-4}$. In addition to visual consequences, the MetaWorld training branches provide short-horizon success and expert-action alignment as training-only supervision.

After WAM training, the three selected members are frozen and the common four-outcome comparison head is trained for up to 40 epochs with learning rate $3\times10^{-4}$ and weight decay $10^{-4}$. Its batch size is 64 anchors on CLIPort and 128 on MetaWorld. The head checkpoint minimizes weighted validation cross-entropy. Fitted feature statistics use the training partition, and task-balanced sampling prevents larger task archives from dominating comparison learning. The original action policies and pretrained visual and language encoders remain frozen throughout these stages. The three WAMs share the observation encoding of a decision state; their latent transitions and the comparison head operate on the resulting features without rendering candidate futures.

\paragraph{Task-chain adaptation.}
The CALVIN instantiation wraps the released FLOWER policy~\cite{reuss2025flower}. Candidate generation exposes five ten-step, seven-dimensional action chunks. Context encoding uses the current image, current language instruction, 15-dimensional observed robot state, and elapsed control progress. The spatial consequence predictor and comparison head retain the same roles as in the CLIPort instantiation, with action and state input layers adapted to these dimensions. The predictive modules use 194 training anchors and 69 validation anchors covering all 34 task labels. Transition supervision describes the candidate's immediate effect, while outcome supervision records completion of the entire remaining task chain. The chain and task-stage indices organize the branch records.

%% file: sections/5.experiments.tex
\section{Experiments}\label{sec:experiments}

We evaluate whether consequence-aware selection improves a frozen policy across different action representations, which parts of predictive learning account for the improvement, and how the result depends on model capacity and representation. MetaWorld tests repeated chunk selection during continuous control; CLIPort provides controlled comparisons of alternative manipulation primitives; CALVIN tests intervention within a sequence of language-conditioned tasks. We report complete-task success, including task-level regressions, and use matched ablations to distinguish the predictive ensemble from the additional consequence-comparison head.

\subsection{Evaluation Protocol}\label{sec:eval-protocol}

\paragraph{Benchmarks and success criterion.}
MetaWorld MT50~\cite{yu2020metaworld} contains 50 manipulation tasks. We use the V3 environments with 50 evaluation episodes per task, for a total of 2,500 episodes. The frozen public $\pi_{0.5}$ policy is the common action backend. Full recomputes five candidate chunks from its current observation at each decision, executes up to eight actions, and observes the resulting state before making the next choice. Success is the environment-defined completion of the task within a 300-step budget. The direct reference uses the same policy and matched task/seed identities.

CLIPort~\cite{shridhar2022cliport} covers 18 task settings, including seen and unseen colors or object categories, with 100 evaluation instances per setting. We measure binary complete-task success over all 1,800 instances. To isolate the effect of candidate selection, each candidate is executed from the same initial state and then followed by the same frozen CLIPort policy until termination. The selector chooses the initial candidate using only the initial context and predicted consequences; its complete-task result is obtained from that candidate's recorded continuation. This protocol measures a single initial intervention. The repeated-selection evaluation is provided by MT50.

\paragraph{Task chains and additional comparisons.}
CALVIN uses 1,000 five-task chains with the released FLOWER policy. We report the fraction completing at least one through five tasks and the average completed chain length. The task-chain protocol is detailed in Section~\ref{sec:calvin-extension}. Comparisons with external policy-augmentation methods use the MT50 and CLIPort evaluation sizes of 2,500 episodes and 1,800 instances, respectively (Section~\ref{sec:external-comparisons}).

\paragraph{Comparisons and reporting.}
\emph{Direct} executes the original policy proposal. \emph{Core} selects with the three-WAM mean-minus-standard-deviation score. \emph{Full} adds the predicted-effect comparison head with fixed residual coefficient~0.25. The candidate set, execution budget, and base policy are shared within each comparison. Future observations and branch outcomes supervise training or score completed decisions; they are never inputs to selection. Evaluation task/seed identities are disjoint from model fitting. We report all task settings with equal weight, so the task-macro average equals the pooled success rate for each benchmark. A \emph{rescue} changes a Direct failure to success, whereas a \emph{harm} changes a Direct success to failure. Their difference equals the net change in successful episodes.

\input{tables/wam_overall}

\begin{figure}[!htb]
\centering
\includegraphics[width=\linewidth]{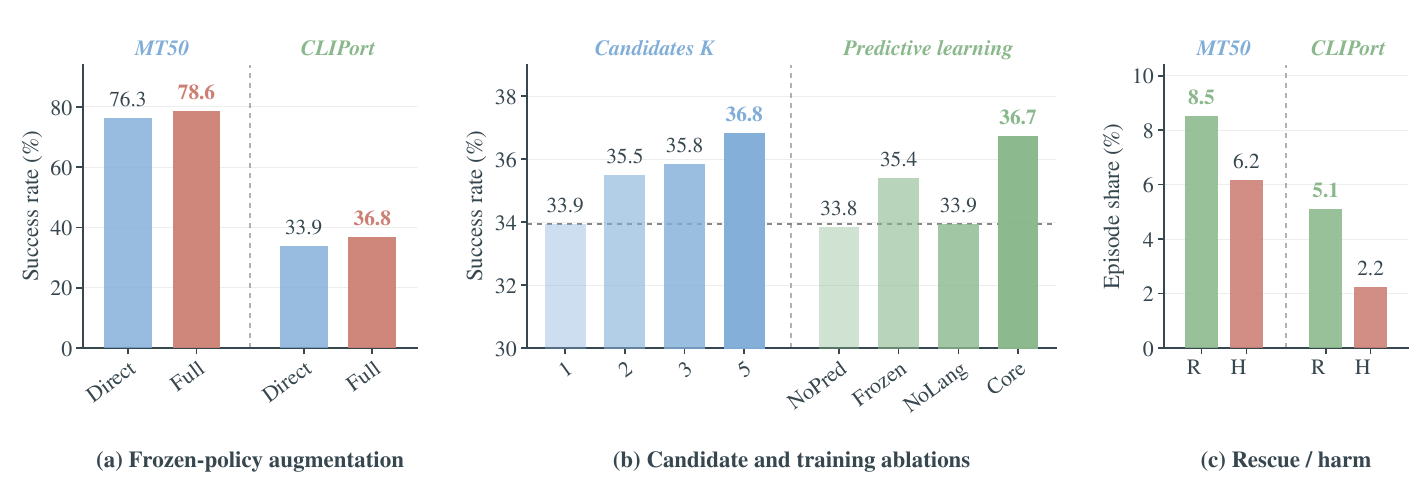}
\caption{Performance and predictive-learning ablations on MT50 and CLIPort. (a) Full improves both frozen policy backends. (b) CLIPort success with different eligible candidate counts and training signals. NoPred removes consequence supervision; Frozen fixes the predictor during outcome learning; NoLang removes language conditioning. (c) Rescue (R) and harm (H), expressed as a percentage of all episodes. The dashed horizontal line in (b) is Direct; its vertical axis is expanded to show differences. Exact values appear in Tables~\ref{tab:cliport-tasks} and~\ref{tab:cliport-objectives}.}
\label{fig:wam-main-results}
\end{figure}

\subsection{Main Results}\label{sec:main-results}

Table~\ref{tab:wam-overall} and Figure~\ref{fig:wam-main-results}(a) summarize the overall results. On MT50, Full increases success from 76.28\% to 78.64\%, adding 59 completed tasks across 2,500 episodes. On CLIPort, success increases from 33.94\% to 36.83\%, adding 52 completed tasks across 1,800 instances. These gains arise without updating the underlying action policies. They cover both stochastic continuous-action proposals and spatial manipulation hypotheses, supporting the use of a shared prediction-and-selection interface across the two backends.

\paragraph{Performance across tasks.}
Table~\ref{tab:metaworld-tasks} reports all 50 MT50 tasks. Full improves 15 tasks, ties Direct on 27, and declines on eight. The largest increases occur on stick push (+40 percentage points), hammer and sweep (+34 each), and basketball (+28). These gains are accompanied by regressions, including dial turn ($-30$ points) and coffee pull ($-22$).

Table~\ref{tab:cliport-tasks} reports all 18 CLIPort settings. Full improves 13 settings, ties on three, and declines on two. Block-in-bowl with unseen colors and seen-color Towers of Hanoi each gain 13 points, while packing shapes gains seven and unseen-color packing boxes gains six. Seen-color block-in-bowl remains at 100\%, whereas unseen-color block pyramid remains at zero. Unseen-color Hanoi drops from 24\% to 17\%.

\input{tables/metaworld_tasks}
\input{tables/cliport_tasks}

\paragraph{Beneficial and harmful interventions.}
The aggregate improvement reflects a balance between successful corrections and disrupted direct trajectories. Full produces 213 rescues and 154 harms on MT50, and 92 rescues and 40 harms on CLIPort (Figure~\ref{fig:wam-main-results}(c)). On CLIPort, it changes the initial choice in 978 of 1,800 instances. Most changes preserve the terminal outcome; the net gain is determined by the 132 instances whose success status changes.

An exact two-sided McNemar test on the paired episode outcomes gives $p=0.0024$ for MT50 and $p<10^{-5}$ for CLIPort. These tests quantify imbalance between rescue and harm on the fixed evaluation identities. A separate task-resampling sensitivity analysis is reported in Section~\ref{sec:limitations}.

\subsection{Comparison with Policy Augmentation Methods}\label{sec:external-comparisons}

\begin{figure}[!htb]
\centering
\includegraphics[width=\linewidth]{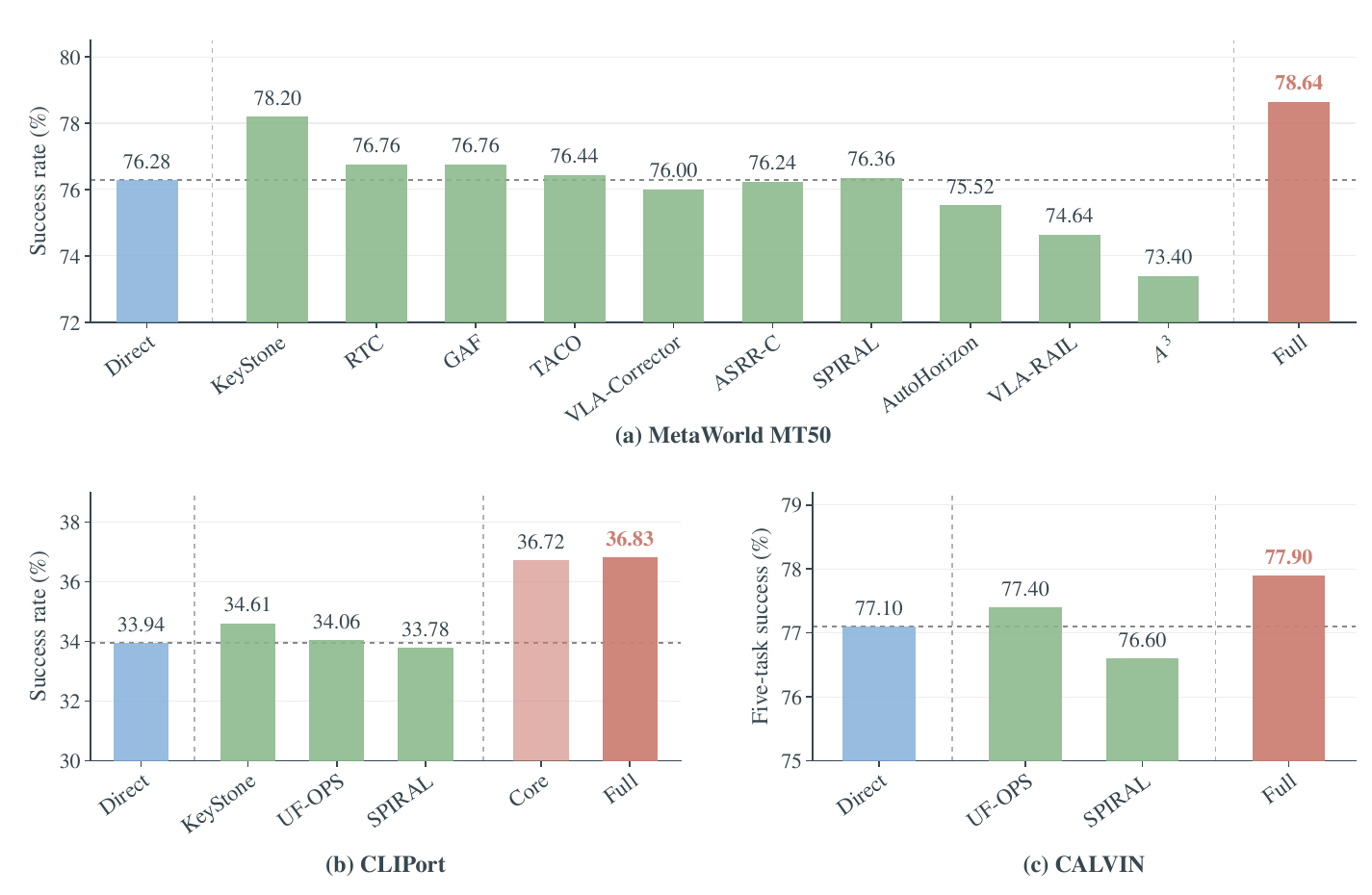}
\caption{Comparison with policy-augmentation methods on (a) MetaWorld MT50, (b) CLIPort, and (c) CALVIN. Panels (a,b) report complete-task success; panel (c) reports five-task-chain success (SR5). Expanded vertical scales show the differences. Blue denotes Direct, green denotes external methods, and brick red denotes EmbodiedSkills. Horizontal dashed lines mark Direct, and vertical dashed lines separate method groups. Numerical values and method adaptations appear in Tables~\ref{tab:policy-augmentation} and~\ref{tab:calvin-chains}.}
\label{fig:policy-augmentation}
\end{figure}

We compare EmbodiedSkills with geometric candidate selection, learned evaluators, action refinement, and adaptive execution methods (Table~\ref{tab:policy-augmentation} and Figure~\ref{fig:policy-augmentation}(a,b)). MetaWorld MT50 includes 50 tasks with 50 evaluation episodes each, and CLIPort includes 18 settings with 100 instances each. Author implementations are adapted to the local frozen policies. UF-OPS and SPIRAL denote the contrastive-verifier and sparse Monte Carlo critic components, respectively. Appendix~\ref{sec:expanded-comparison-protocols} specifies the adaptations and execution schedules.

\input{tables/policy_augmentation_comparison}

On MT50, Full completes 1,966 of 2,500 episodes (78.64\%), compared with 1,907 for Direct (76.28\%) and 1,955 for KeyStone (78.20\%). RTC and GAF each complete 1,919 episodes, while TACO completes 1,911. Full therefore improves on Direct by 2.36 percentage points and on the strongest external comparator, KeyStone, by 0.44 points. The smaller margin over geometric self-consistency indicates that action-space agreement already captures useful information; predicted consequences provide a further refinement. RTC executes four actions with a simulated two-step delay, whereas Direct and Full execute up to eight without simulated delay.

On CLIPort, Full completes 663 of 1,800 tasks (36.83\%), compared with 611 for Direct (33.94\%), 623 for KeyStone (34.61\%), 613 for the UF-OPS verifier (34.06\%), and 608 for the SPIRAL critic (33.78\%). The gains over the three external alternatives are 2.22, 2.78, and 3.06 percentage points, respectively. Core completes 661 tasks (36.72\%), preserving most of Full's improvement. These comparisons support candidate-conditioned consequence prediction alongside the geometric and direct outcome-scoring alternatives, with the extra comparison head providing a small final correction.

\subsection{Long-Horizon Adaptation on CALVIN}\label{sec:calvin-extension}

CALVIN~\cite{mees2022calvinbenchmarklanguageconditionedpolicy} evaluates sequential execution of language-conditioned manipulation tasks. We use the released FLOWER policy~\cite{reuss2025flower} on 1,000 five-task chains, with a 360-control-step budget per task. At a predetermined point in one assigned task, the frozen policy proposes five ten-step chunks. Each candidate is followed by the same frozen policy through the remaining chain. Full, the UF-OPS verifier, and the SPIRAL critic select using the current observation, current instruction, observed robot state, and candidate actions. Direct executes candidate~0 with the same chunk-boundary policy reset. The comparison measures one intervention within a chain, with complete-chain success as the outcome.

\input{tables/calvin_chain_comparison}

As shown in Figure~\ref{fig:policy-augmentation}(c), Full increases complete-chain success from 771/1,000 to 779/1,000 (77.10\% to 77.90\%) and average completed length from 4.509 to 4.522. The UF-OPS verifier completes 774 chains with mean length 4.515, and the SPIRAL critic completes 766 with mean length 4.496. Relative to Direct, Full matches SR1 and SR3, improves SR2 by 0.10 percentage points, SR4 by 0.40 points, and SR5 by 0.80 points. The larger difference at the end of the chain is consistent with the complete-chain training objective. This extends consequence-aware selection to a third frozen action backend through the shared skill interfaces.

SR$k$ measures the fraction completing at least $k$ consecutive tasks from the beginning of the chain. All 1,000 chains contribute to the denominator, including those that terminate before the intervention point. Average completed length sums the five completion fractions, so it captures partial progress as well as completion of the entire sequence. Appendix~\ref{sec:calvin-chain-protocol} describes the chain protocol and the distinction between immediate consequence prediction and full-chain outcome supervision.

\FloatBarrier

\subsection{Candidate Selection and Predictive Learning}\label{sec:mechanism-ablation}

\paragraph{Candidate coverage.}
Restricting Full to the first $K\in\{1,2,3,5\}$ candidates gives 33.94\%, 35.50\%, 35.83\%, and 36.83\% success, respectively (Table~\ref{tab:cliport-tasks}). This decision-set coverage study masks the eligible prefix of an already scored five-candidate set. More alternatives provide additional opportunities to recover unsuccessful direct actions.

\begin{figure}[!htb]
\centering
\includegraphics[width=\linewidth]{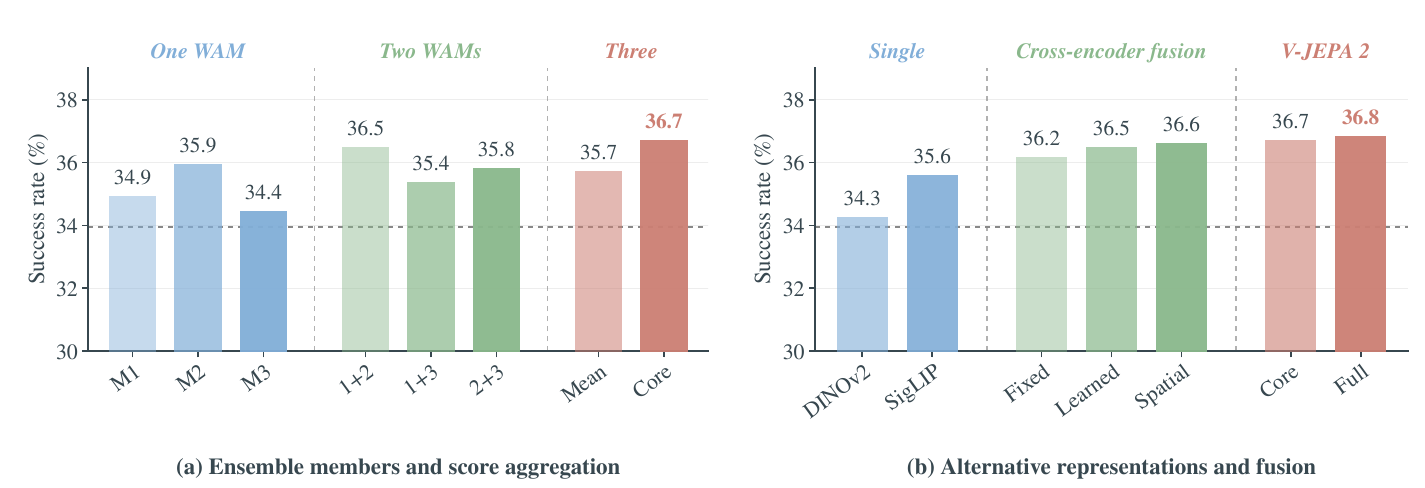}
\caption{Model comparisons on CLIPort. (a) All three single WAMs, all three two-member combinations, and the two three-member aggregation rules. Two-member rules and Core subtract score standard deviation; Mean uses only the arithmetic mean. (b) Individual visual representations and cross-encoder fusion systems. Fixed and Learned fuse V-JEPA~2 with DINOv2; Spatial fuses V-JEPA~2 and SigLIP. These systems differ in model multiplicity and fusion structure. Dashed horizontal lines indicate Direct; expanded vertical axes show the measured differences.}
\label{fig:wam-model-comparisons}
\end{figure}

\paragraph{Independent predictions and disagreement.}
Figure~\ref{fig:wam-model-comparisons}(a) compares every individual WAM and all three two-member combinations. Single members achieve 34.94\%, 35.94\%, and 34.44\%; the two-member rules achieve 36.50\%, 35.39\%, and 35.83\%. The arithmetic mean of all three reaches 35.72\%, while subtracting their score standard deviation raises success to 36.72\%. Relative to the arithmetic mean, Core reduces harms from 71 to 37 while rescues decrease from 103 to 87, yielding 18 additional successes overall. The gain comes from a better balance of corrections and mistakes. Table~\ref{tab:cliport-ensembles} provides every task-level result.

\paragraph{Does predicting consequences matter?}
Table~\ref{tab:cliport-objectives} isolates the training signals against Core. Removing future-consequence supervision reduces success from 36.72\% to 33.83\%, slightly below Direct. Freezing the predictive WAM during outcome learning retains 35.39\%, but falls 1.33 points below joint learning. Removing language conditioning yields 33.94\%, equal to Direct in aggregate but with 59 rescues and 59 harms. These comparisons support three complementary roles: transition supervision constrains what the candidate is expected to change, outcome gradients adapt that prediction for selection, and language conditions whether the predicted change serves the task.

\input{tables/cliport_objectives}

\paragraph{Contribution of consequence comparison.}
Adding the comparison head increases Core from 661 to 663 successes, a gain of 0.11 percentage points. Full adds five rescues and three harms relative to Core's aggregate accounting against Direct. The predictive Core accounts for most of the measured gain, with consequence comparison providing a small additional refinement.

\subsection{Representations, Capacity, and Optimization}\label{sec:sensitivity}

\paragraph{Visual representations and modular integration.}
Table~\ref{tab:cliport-backbones} and Figure~\ref{fig:wam-model-comparisons}(b) compare alternative consequence systems with different member counts and fusion designs within the same candidate-selection workflow. The individual DINOv2 and SigLIP systems obtain 34.28\% and 35.61\%, respectively. Fixed and learned V-JEPA~2/DINOv2 fusion reach 36.17\% and 36.50\%; spatial V-JEPA~2/SigLIP fusion reaches 36.61\%. Core and Full obtain 36.72\% and 36.83\%. Here the context-encoding and consequence-prediction skills admit alternative representations, while candidate generation and action execution retain the same CLIPort backend. These results demonstrate component substitution within the skill composition defined in Table~\ref{tab:stage-skills}.

\input{tables/cliport_backbones}

\begin{figure}[!htb]
\centering
\includegraphics[width=\linewidth]{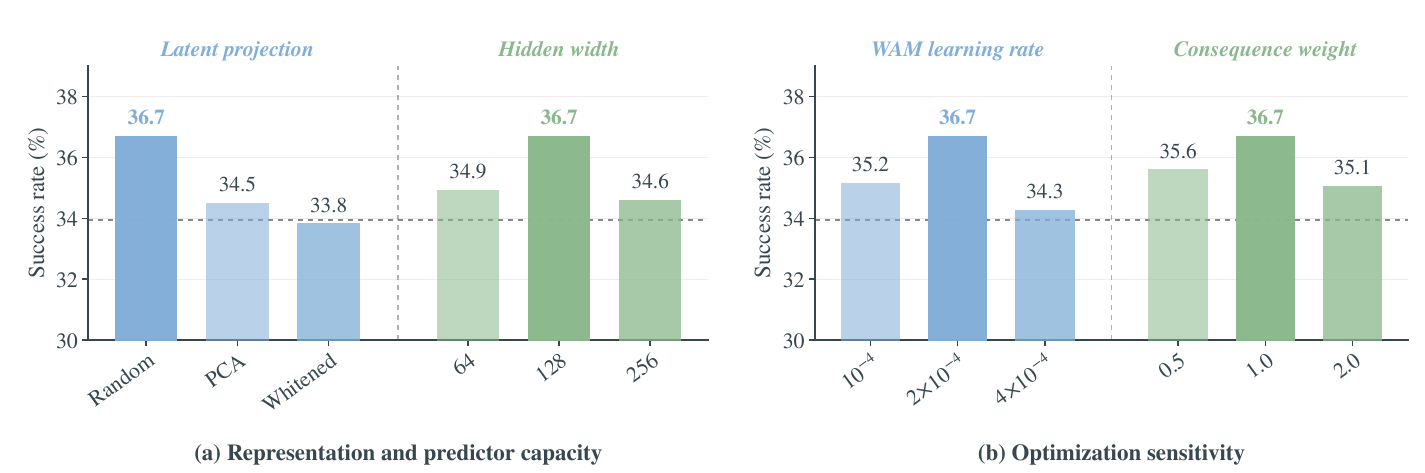}
\caption{One-factor sensitivity studies on CLIPort. (a) Projected visual changes and predictor hidden width. (b) Joint-stage WAM learning rate and consequence-loss coefficient. The default settings are random orthogonal projection, width 128, learning rate $2\times10^{-4}$, and consequence weight 1.0. Each default reuses the same Core result. Dashed horizontal lines mark Direct; the vertical axes are expanded.}
\label{fig:wam-sensitivity}
\end{figure}

\paragraph{Latent projection and predictor capacity.}
Figure~\ref{fig:wam-sensitivity}(a) examines two representation choices relative to Core. A fixed random orthogonal 32-dimensional projection reaches 36.72\%; training-fitted PCA and whitened PCA reach 34.50\% and 33.83\%. Preserving directions of large visual variation is therefore not sufficient to preserve the distinctions useful for selecting actions. Increasing predictor width also does not monotonically improve performance: widths 64, 128, and 256 achieve 34.94\%, 36.72\%, and 34.61\%, respectively. The corresponding transition predictors contain 172,448, 351,008, and 732,704 parameters per member. The intermediate width gives the best result among the tested capacities.

\paragraph{Learning rate and predictive-loss balance.}
Figure~\ref{fig:wam-sensitivity}(b) varies one factor at a time. Predictor learning rates $10^{-4}$, $2\times10^{-4}$, and $4\times10^{-4}$ give 35.17\%, 36.72\%, and 34.28\% success with the value-head learning rate fixed. Consequence-loss weights 0.5, 1.0, and 2.0 give 35.61\%, 36.72\%, and 35.06\%. The default balances learning the physical transition with learning which transitions help the task. Complete task-level sensitivity results appear in Tables~\ref{tab:cliport-representation} and~\ref{tab:cliport-optimization}; Table~\ref{tab:cliport-all-outcomes} lists rescue, harm, and intervention counts for all reported configurations.

\paragraph{Decision-module cost.}
The three CLIPort WAM members and consequence-comparison head contain 9.37 million parameters in total. With five candidate inputs and batch size one, their combined forward pass takes 11.33\,ms on an A100-80GB (100 synchronized iterations after 20 warm-up iterations; 13.28\,ms at the 95th percentile). The timing covers the Full decision module with precomputed observation features and candidate inputs.

\subsection{Qualitative Analysis}\label{sec:qualitative-wam}

\begin{figure}[!htb]
\centering
\includegraphics[width=\linewidth]{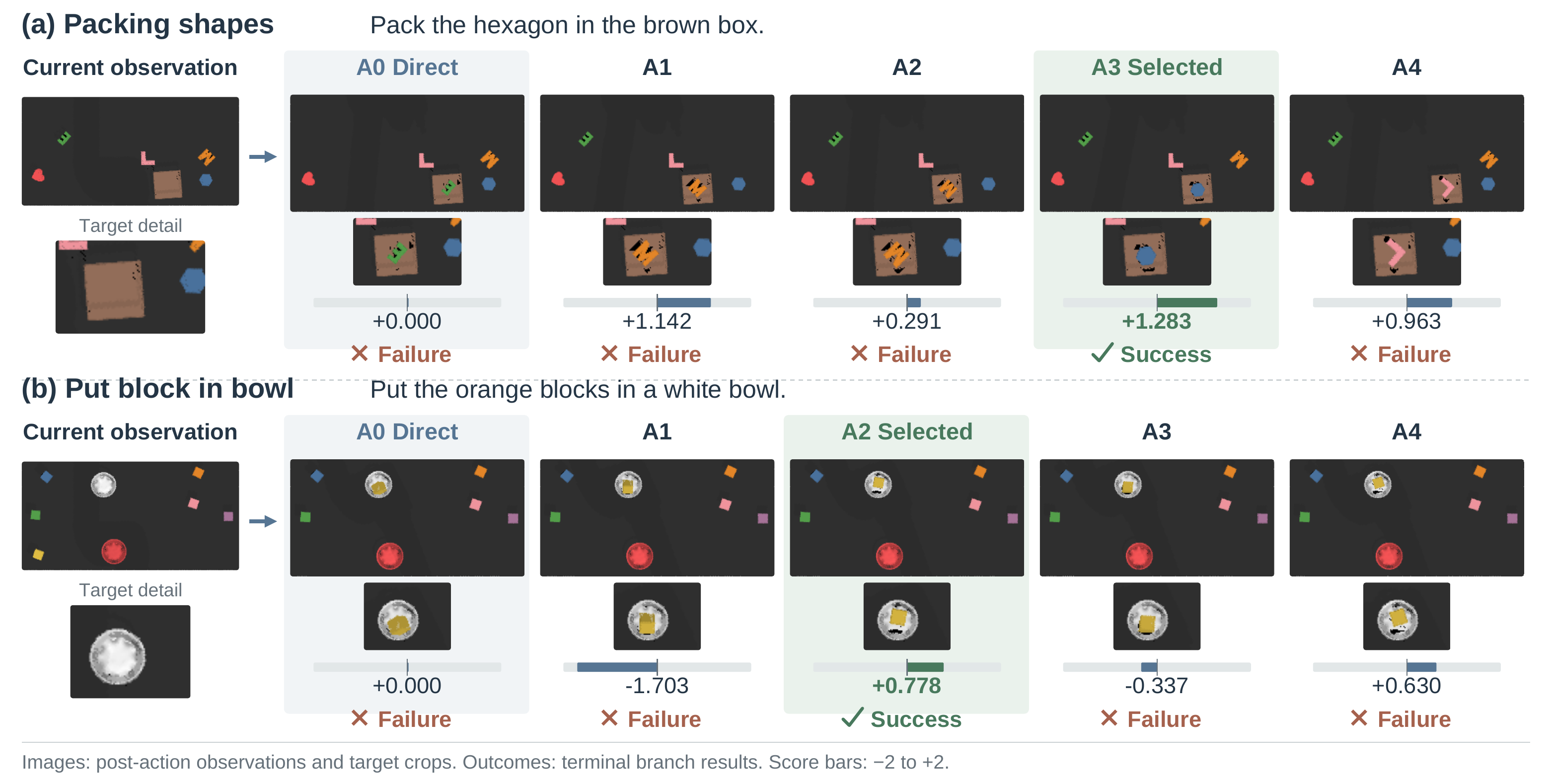}
\caption{Five candidate actions from the same observation in two CLIPort examples. Images show the recorded immediate post-action state and a common target-region crop. Bars show relative selector scores on a shared scale. Terminal success labels refer to each branch after frozen-policy continuation, and are shown for analysis only.}
\label{fig:wam-candidates}
\end{figure}

\paragraph{Comparing plausible actions.}
Figure~\ref{fig:wam-candidates} shows five alternatives from the same observation in two selected successful corrections: packing shapes and placing a block in a bowl. All images are recorded immediate post-action observations, while the success annotations describe the completed branch under the frozen continuation policy. The selected candidate succeeds in both examples and the direct candidate fails. The common target-region crops expose the spatial differences without changing camera scale across candidates.

\begin{figure}[!htb]
\centering
\includegraphics[width=\linewidth]{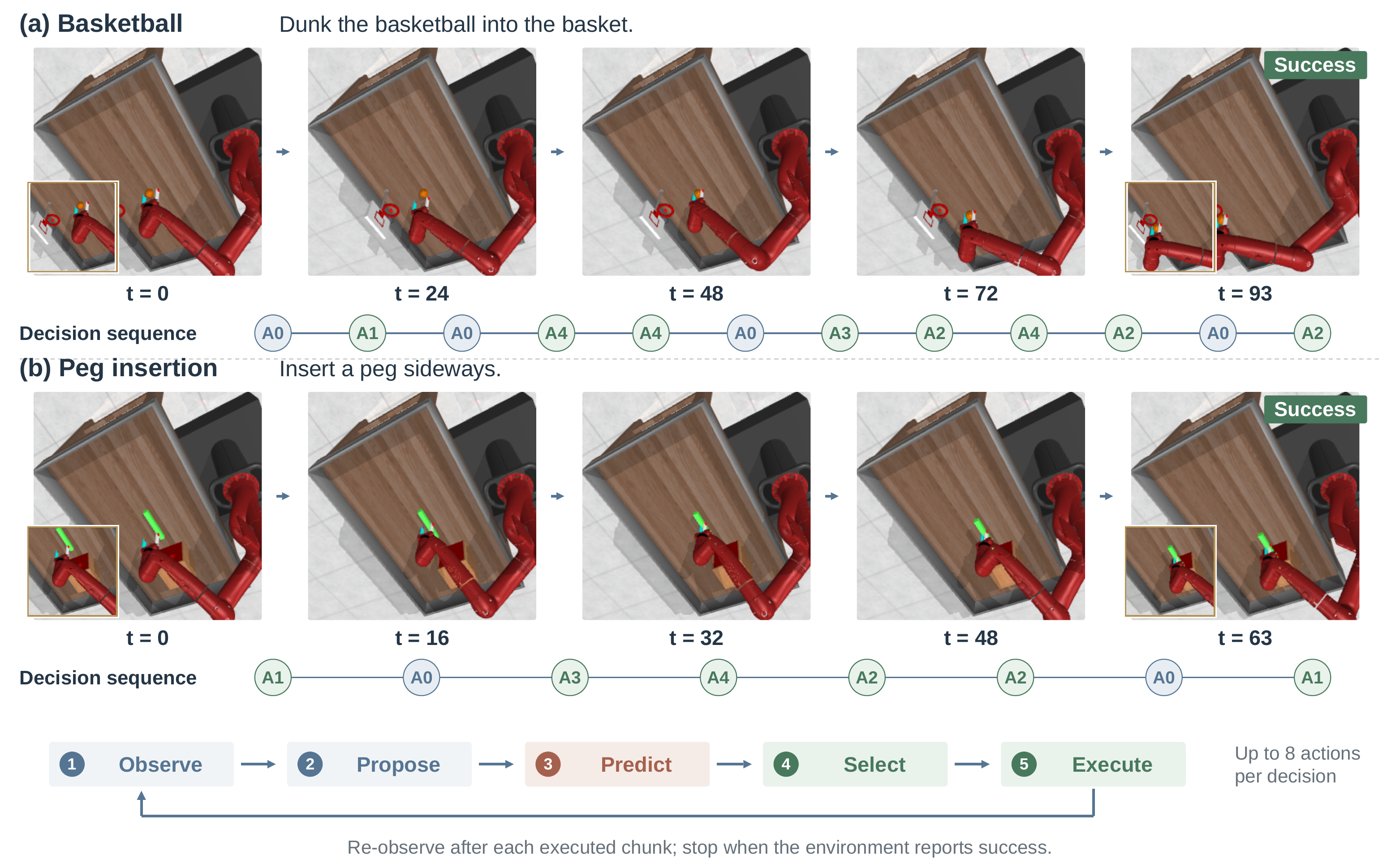}
\caption{Successful closed-loop MetaWorld executions. Keyframes show basketball and peg insertion, while the decision strip records every selected candidate. Each decision executes up to eight low-level actions before receiving a new observation. Insets magnify the corresponding frame. Intermediate images are obtained by replaying the recorded actions, with initial and final observations checked against the source trajectories.}
\label{fig:wam-execution}
\end{figure}

\paragraph{Repeated selection under live feedback.}
Figure~\ref{fig:wam-execution} presents successful MetaWorld basketball and peg-insertion trajectories. Every decision uses the observation reached by the previously selected chunk. The decision strip includes all choices, including those between the displayed keyframes, and shows that the direct candidate remains available throughout execution. The trajectories complete in 93 and 63 low-level steps, respectively. The last action block can be shorter than eight steps when the task terminates. The five-stage loop updates each decision using feedback from the preceding execution.

\begin{figure}[!htb]
\centering
\includegraphics[width=\linewidth]{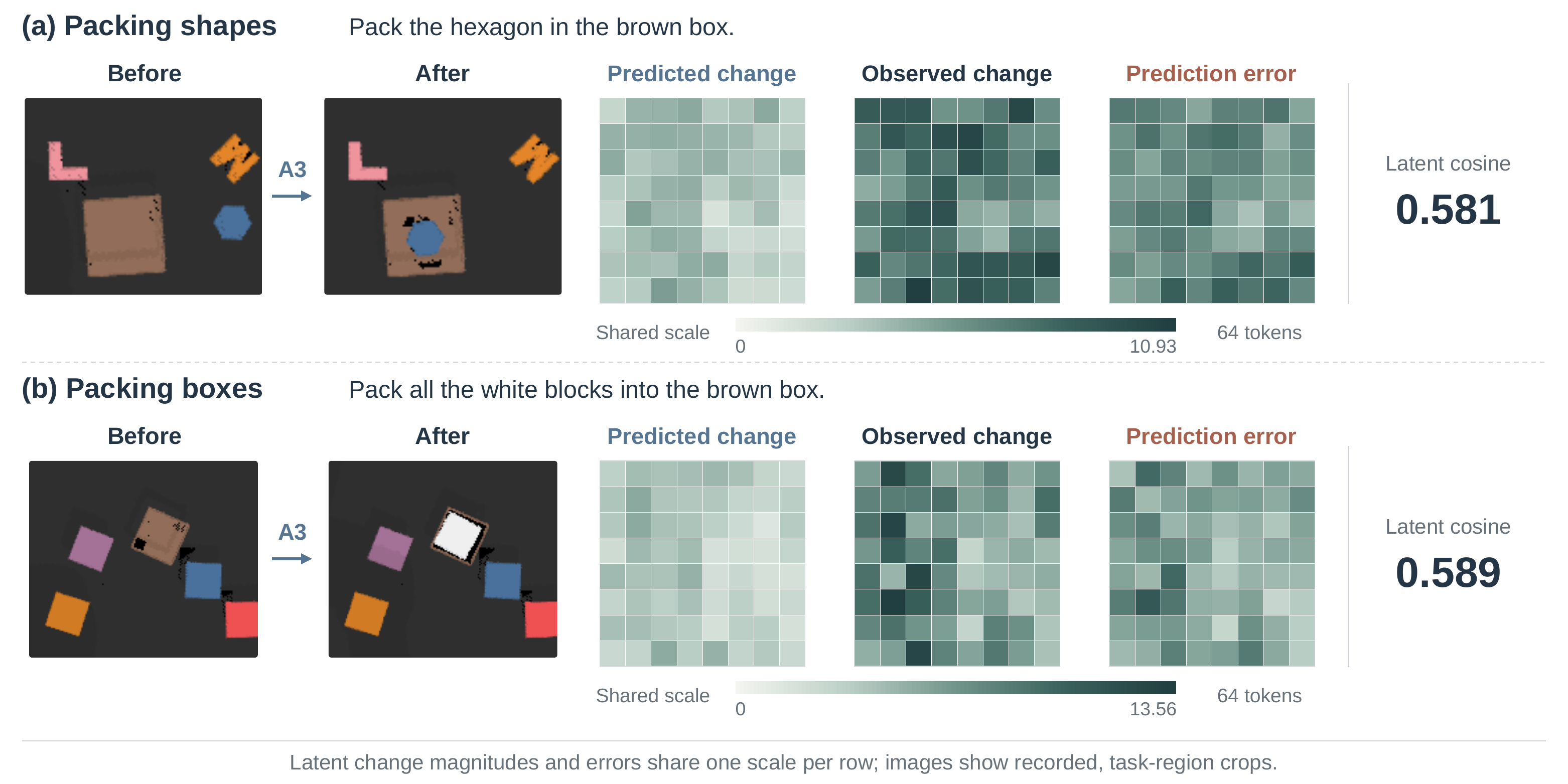}
\caption{Predicted and observed latent consequences for candidate A3 in two held-out CLIPort transitions. Heatmaps show the $L_2$ magnitude of each 32-dimensional projected change at 64 spatial tokens; error is the $L_2$ difference between predicted and observed changes. Prediction averages the three WAM outputs. Each example uses one color scale, and cosine similarity is computed over the flattened latent changes. RGB panels show recorded task-region crops.}
\label{fig:wam-consequences}
\end{figure}

\begin{figure}[!htb]
\centering
\includegraphics[width=\linewidth]{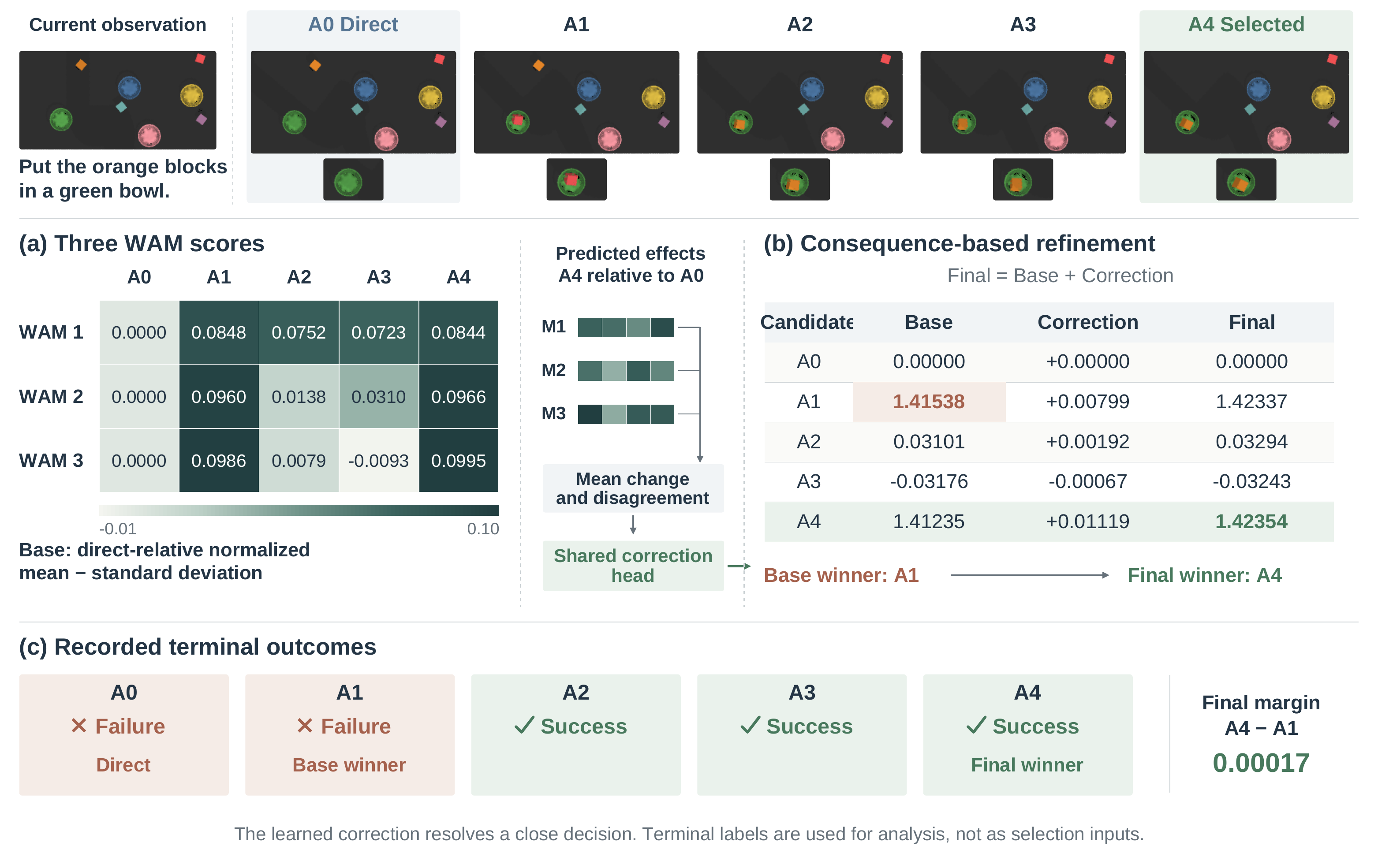}
\caption{A consequence-comparison decision trace. The base ensemble favors A1, while the predicted-effect correction selects A4. Scores are rounded to five decimal places; the final A4--A1 margin is 0.00017. Center tiles summarize groups of A4's standardized candidate-relative predicted effect. All five terminal outcomes appear below for retrospective comparison.}
\label{fig:wam-trace}
\end{figure}

\paragraph{Predicted changes and decision refinement.}
Figure~\ref{fig:wam-consequences} compares predicted latent changes with observed changes for two held-out transitions. The heatmaps show the magnitude of a 32-dimensional projected change at each spatial token, using a common scale within each example. They visualize the spatial distribution of predicted change and residual error. Figure~\ref{fig:wam-trace} then exposes a decision in which the consequence correction changes the ensemble preference from candidate A1 to A4. The final score margin is only 0.00017, so the example illustrates a near-tie refinement. All five terminal outcomes are displayed; candidates A2, A3, and A4 succeed. The small margin and the modest aggregate Core-to-Full gain are consistent with the head acting as a correction to the main predictive score.

%% file: tables/wam_overall.tex
\begin{table}[!htb]
\centering
\small
\setlength{\tabcolsep}{6pt}
\renewcommand{\arraystretch}{1.10}
\begin{tabularx}{\linewidth}{@{}>{\raggedright\arraybackslash}X>{\raggedright\arraybackslash}XrPOA@{}}
\toprule
\rowcolor{wfheader} \textbf{Benchmark} & \textbf{Policy} & \textbf{Episodes} & \textbf{Direct (\%)} & \textbf{Full (\%)} & \textbf{$\Delta$ (pp)} \\
\midrule
MetaWorld MT50 & $\pi_{0.5}$ & 2,500 & 76.28 & \best{78.64} & \gain{+2.36} \\
CLIPort & CLIPort & 1,800 & 33.94 & \best{36.83} & \gain{+2.89} \\
CALVIN (SR5) & FLOWER & 1,000 & 77.10 & \best{77.90} & \gain{+0.80} \\
\bottomrule
\end{tabularx}
\caption{Complete-task success with each frozen action backend. MT50 uses repeated selection; CLIPort uses one initial intervention; CALVIN makes one intervention within a five-task chain and reports completion of all five tasks. The CALVIN chain protocol and additional metrics appear in Section~\ref{sec:calvin-extension}.}
\label{tab:wam-overall}
\end{table}

%% file: tables/metaworld_tasks.tex
\begin{table}[!htb]
\centering
\small
\setlength{\tabcolsep}{3.3pt}
\renewcommand{\arraystretch}{1.10}
\begin{tabularx}{\linewidth}{@{}>{\raggedright\arraybackslash}XPOA>{\raggedright\arraybackslash}XPOA@{}}
\toprule
\rowcolor{wfheader} \textbf{Task} & \textbf{Direct} & \textbf{Full} & \textbf{$\Delta$} & \textbf{Task} & \textbf{Direct} & \textbf{Full} & \textbf{$\Delta$} \\
\midrule
Assembly & 92 & \textbf{100} & \gain{+8} & Handle pull (side) & 0 & \textbf{14} & \gain{+14} \\
Basketball & 52 & \textbf{80} & \gain{+28} & Handle pull & 0 & \textbf{24} & \gain{+24} \\
Bin picking & 44 & \textbf{52} & \gain{+8} & Lever pull & 42 & \textbf{58} & \gain{+16} \\
Box close & \textbf{60} & \textbf{60} & \same{0} & Peg insert (side) & 78 & \textbf{88} & \gain{+10} \\
Button topdown & \textbf{100} & \textbf{100} & \same{0} & Peg unplug (side) & \textbf{100} & 82 & \loss{-18} \\
Button topdown (wall) & \textbf{100} & \textbf{100} & \same{0} & Pick out of hole & \textbf{42} & 26 & \loss{-16} \\
Button press & \textbf{100} & \textbf{100} & \same{0} & Pick place & \textbf{56} & \textbf{56} & \same{0} \\
Button press (wall) & \textbf{100} & \textbf{100} & \same{0} & Pick place (wall) & \textbf{94} & \textbf{94} & \same{0} \\
Coffee button & \textbf{100} & \textbf{100} & \same{0} & Plate slide back (side) & \textbf{100} & 88 & \loss{-12} \\
Coffee pull & \textbf{66} & 44 & \loss{-22} & Plate slide back & \textbf{92} & \textbf{92} & \same{0} \\
Coffee push & \textbf{100} & \textbf{100} & \same{0} & Plate slide (side) & \textbf{100} & \textbf{100} & \same{0} \\
Dial turn & \textbf{48} & 18 & \loss{-30} & Plate slide & \textbf{100} & \textbf{100} & \same{0} \\
Disassemble & \textbf{72} & \textbf{72} & \same{0} & Push back & 42 & \textbf{48} & \gain{+6} \\
Door close & 84 & \textbf{100} & \gain{+16} & Push & \textbf{100} & \textbf{100} & \same{0} \\
Door lock & 82 & \textbf{94} & \gain{+12} & Push wall & \textbf{100} & \textbf{100} & \same{0} \\
Door open & \textbf{100} & \textbf{100} & \same{0} & Reach & \textbf{30} & \textbf{30} & \same{0} \\
Door unlock & \textbf{100} & 86 & \loss{-14} & Reach wall & \textbf{82} & \textbf{82} & \same{0} \\
Drawer close & \textbf{100} & \textbf{100} & \same{0} & Shelf place & \textbf{54} & \textbf{54} & \same{0} \\
Drawer open & \textbf{100} & \textbf{100} & \same{0} & Soccer & \textbf{52} & 36 & \loss{-16} \\
Faucet close & 84 & \textbf{88} & \gain{+4} & Stick pull & \textbf{38} & 18 & \loss{-20} \\
Faucet open & \textbf{100} & \textbf{100} & \same{0} & Stick push & 50 & \textbf{90} & \gain{+40} \\
Hammer & 48 & \textbf{82} & \gain{+34} & Sweep into & 78 & \textbf{90} & \gain{+12} \\
Hand insert & \textbf{86} & \textbf{86} & \same{0} & Sweep & 66 & \textbf{100} & \gain{+34} \\
Handle press (side) & \textbf{100} & \textbf{100} & \same{0} & Window close & \textbf{100} & \textbf{100} & \same{0} \\
Handle press & \textbf{100} & \textbf{100} & \same{0} & Window open & \textbf{100} & \textbf{100} & \same{0} \\
\midrule
\rowcolor{wfaveragerow} \textbf{Average (\%)} & 76.28 & \textbf{78.64} & \gain{+2.36} & \textbf{Successes / 2,500} & 1,907 & \textbf{1,966} & \gain{+59} \\
\bottomrule
\end{tabularx}
\caption{Success rates (\%) on all 50 MetaWorld V3 tasks, with 50 episodes per task. Direct and Full use the same frozen $\pi_{0.5}$ policy; Full selects among five proposed chunks at each decision. $\Delta$ is in percentage points; one success corresponds to two points within a task. Best results in each task are bold.}
\label{tab:metaworld-tasks}
\end{table}

%% file: tables/cliport_tasks.tex
\begin{table}[!htb]
\centering
\small
\setlength{\tabcolsep}{5.5pt}
\renewcommand{\arraystretch}{1.10}
\begin{tabularx}{\linewidth}{@{}>{\raggedright\arraybackslash}XPrrQOA@{}}
\toprule
\rowcolor{wfheader} \textbf{Task setting} & \textbf{Direct} & \textbf{$K{=}2$} & \textbf{$K{=}3$} & \textbf{Core} & \textbf{Full} & \textbf{$\Delta$} \\
\midrule
Align rope & 7 & 8 & 7 & \textbf{10} & 8 & \gain{+1} \\
Assembling kits (seen) & 20 & 18 & 23 & \textbf{24} & 22 & \gain{+2} \\
Assembling kits (unseen) & 2 & 2 & 0 & 3 & \textbf{4} & \gain{+2} \\
Packing boxes (seen) & 50 & 50 & 50 & \textbf{54} & 52 & \gain{+2} \\
Packing boxes (unseen) & 14 & 18 & \textbf{20} & \textbf{20} & \textbf{20} & \gain{+6} \\
Google objects, group (seen) & 21 & 24 & \textbf{26} & 22 & 24 & \gain{+3} \\
Google objects, seq. (seen) & 17 & 16 & 19 & \textbf{22} & 18 & \gain{+1} \\
Packing shapes & 36 & 40 & 41 & \textbf{44} & 43 & \gain{+7} \\
Google objects, group (unseen) & 20 & 19 & 15 & 20 & \textbf{22} & \gain{+2} \\
Google objects, seq. (unseen) & \textbf{7} & 6 & 4 & 3 & 5 & \loss{-2} \\
Block in bowl (seen) & \textbf{100} & \textbf{100} & \textbf{100} & \textbf{100} & \textbf{100} & \same{0} \\
Block in bowl (unseen) & 37 & 41 & 49 & 46 & \textbf{50} & \gain{+13} \\
Separating piles (seen) & \textbf{64} & 60 & 60 & 63 & \textbf{64} & \same{0} \\
Separating piles (unseen) & 41 & 42 & 42 & 42 & \textbf{44} & \gain{+3} \\
Block pyramid (seen) & 80 & \textbf{87} & 84 & 86 & 86 & \gain{+6} \\
Block pyramid (unseen) & \textbf{0} & \textbf{0} & \textbf{0} & \textbf{0} & \textbf{0} & \same{0} \\
Towers of Hanoi (seen) & 71 & 81 & 82 & \textbf{84} & \textbf{84} & \gain{+13} \\
Towers of Hanoi (unseen) & 24 & \textbf{27} & 23 & 18 & 17 & \loss{-7} \\
\midrule
\rowcolor{wftotalrow} \textbf{Successes / 1,800} & 611 & 639 & 645 & 661 & \textbf{663} & \gain{+52} \\
\rowcolor{wfaveragerow} \textbf{Macro success (\%)} & 33.94 & 35.50 & 35.83 & 36.72 & \textbf{36.83} & \gain{+2.89} \\
\bottomrule
\end{tabularx}
\caption{Complete-task success rates (\%) on all 18 CLIPort settings, with 100 trials per setting. Direct is candidate 0; Core uses three WAMs with mean-minus-standard-deviation aggregation; Full adds consequence comparison. The $K=2,3$ columns restrict the eligible prefix of the Full candidate set. $\Delta$ compares Full with Direct. Seen/unseen follows the task setting name; seq. denotes sequential instructions. Best success values in each row are bold, including ties.}
\label{tab:cliport-tasks}
\end{table}

%% file: tables/policy_augmentation_comparison.tex
\begin{table}[!htb]
\centering
\small
\setlength{\tabcolsep}{4pt}
\renewcommand{\arraystretch}{1.12}
\begin{tabularx}{\linewidth}{@{}>{\raggedright\arraybackslash}Xrrrr@{}}
\toprule
\rowcolor{wfheader} \textbf{Method / adaptation} & \textbf{MT50 / 2,500} & \textbf{\%} & \textbf{CLIPort / 1,800} & \textbf{\%} \\
\midrule
\rowcolor{wfapi} Direct & 1,907 & 76.28 & 611 & 33.94 \\
KeyStone~\cite{dai2026geometryguidedselfconsistencyphysical} & 1,955 & 78.20 & 623 & 34.61 \\
RTC (E4, delay 2)~\cite{black2025realtimeexecutionactionchunking} & 1,919 & 76.76 & \na & \na \\
GAF (early implementation)~\cite{yang2026guidedactionflowvalueguided} & 1,919 & 76.76 & \na & \na \\
TACO (CFN)~\cite{yang2025taco} & 1,911 & 76.44 & \na & \na \\
VLA-Corrector (8-step)~\cite{pan2026vlacorrector} & 1,900 & 76.00 & \na & \na \\
ASRR-C (width 256)~\cite{asrr2026} & 1,906 & 76.24 & \na & \na \\
SPIRAL (MC critic)~\cite{chen2026sarm2multitaskstageaware} & 1,909 & 76.36 & 608 & 33.78 \\
AutoHorizon~\cite{wang2026vlaknowslimitsadaptive} & 1,888 & 75.52 & \na & \na \\
VLA-RAIL (synchronous)~\cite{zhao2025vlarailrealtimeasynchronousinference} & 1,866 & 74.64 & \na & \na \\
A$^3$~\cite{chen2026dynamicexecutioncommitmentvisionlanguageaction} & 1,835 & 73.40 & \na & \na \\
UF-OPS (verifier)~\cite{attarian2026updatefreeonpolicysteeringverifiers} & \na & \na & 613 & 34.06 \\
\midrule
\rowcolor{wftotalrow} EmbodiedSkills Core & \na & \na & 661 & 36.72 \\
\rowcolor{wfaveragerow} \textbf{EmbodiedSkills Full} & \textbf{1,966} & \textbf{78.64} & \textbf{663} & \textbf{36.83} \\
\bottomrule
\end{tabularx}
\caption{Policy-augmentation comparisons on MetaWorld MT50 (50 tasks, 50 episodes per task) and CLIPort (18 settings, 100 instances per setting). External methods use author implementations or released components adapted to the local frozen policy. RTC uses its stated execution schedule; UF-OPS and SPIRAL denote verifier and MC-critic adaptations. Dashes indicate untested combinations. Best values within each benchmark are bold. Appendix~\ref{sec:expanded-comparison-protocols} details the adaptations.}
\label{tab:policy-augmentation}
\end{table}

%% file: tables/calvin_chain_comparison.tex
\begin{table}[!htb]
\centering
\small
\setlength{\tabcolsep}{4pt}
\renewcommand{\arraystretch}{1.12}
\begin{tabularx}{\linewidth}{@{}>{\raggedright\arraybackslash}Xrrrrrr@{}}
\toprule
\rowcolor{wfheader} \textbf{Method} & \textbf{SR1} & \textbf{SR2} & \textbf{SR3} & \textbf{SR4} & \textbf{SR5} & \textbf{Avg. length} \\
\midrule
\rowcolor{wfapi} Direct FLOWER~\cite{reuss2025flower} & 99.50 & 96.20 & \textbf{91.70} & 86.40 & 77.10 & 4.509 \\
UF-OPS (verifier)~\cite{attarian2026updatefreeonpolicysteeringverifiers} & \textbf{99.80} & \textbf{96.30} & 91.40 & 86.60 & 77.40 & 4.515 \\
SPIRAL (MC critic)~\cite{chen2026sarm2multitaskstageaware} & 99.50 & 95.80 & 91.50 & 86.20 & 76.60 & 4.496 \\
\rowcolor{wfaveragerow} \textbf{EmbodiedSkills Full} & 99.50 & \textbf{96.30} & \textbf{91.70} & \textbf{86.80} & \textbf{77.90} & \textbf{4.522} \\
\bottomrule
\end{tabularx}
\caption{CALVIN evaluation with the frozen FLOWER policy over 1,000 five-task chains. SR$k$ is the percentage completing at least $k$ consecutive tasks, and average completed length equals $\sum_{k=1}^{5}\mathrm{SR}k/100$. Direct, UF-OPS, SPIRAL, and Full complete all five tasks in 771, 774, 766, and 779 chains, respectively. Best values in each column are bold.}
\label{tab:calvin-chains}
\end{table}

%% file: tables/cliport_objectives.tex
\begin{table}[!htb]
\centering
\small
\setlength{\tabcolsep}{7pt}
\renewcommand{\arraystretch}{1.10}
\begin{tabularx}{\linewidth}{@{}>{\raggedright\arraybackslash}XPOrrr@{}}
\toprule
\rowcolor{wfheader} \textbf{Task setting} & \textbf{Direct} & \textbf{Core} & \textbf{NoPred} & \textbf{Frozen} & \textbf{NoLang} \\
\midrule
Align rope & 7 & \textbf{10} & 5 & \textbf{10} & 2 \\
Assembling kits (seen) & 20 & \textbf{24} & 15 & 16 & 14 \\
Assembling kits (unseen) & 2 & \textbf{3} & 0 & 1 & 0 \\
Packing boxes (seen) & 50 & \textbf{54} & 50 & 52 & 50 \\
Packing boxes (unseen) & 14 & 20 & 20 & \textbf{22} & 18 \\
Google objects, group (seen) & 21 & 22 & 24 & \textbf{28} & 24 \\
Google objects, seq. (seen) & 17 & 22 & \textbf{24} & 19 & 20 \\
Packing shapes & 36 & \textbf{44} & 32 & 32 & 36 \\
Google objects, group (unseen) & 20 & 20 & 15 & 18 & \textbf{21} \\
Google objects, seq. (unseen) & 7 & 3 & 0 & \textbf{10} & 4 \\
Block in bowl (seen) & \textbf{100} & \textbf{100} & \textbf{100} & \textbf{100} & 96 \\
Block in bowl (unseen) & 37 & 46 & 48 & 41 & \textbf{50} \\
Separating piles (seen) & \textbf{64} & 63 & 57 & 62 & \textbf{64} \\
Separating piles (unseen) & 41 & 42 & \textbf{44} & 42 & 37 \\
Block pyramid (seen) & 80 & \textbf{86} & 81 & \textbf{86} & 78 \\
Block pyramid (unseen) & \textbf{0} & \textbf{0} & \textbf{0} & \textbf{0} & \textbf{0} \\
Towers of Hanoi (seen) & 71 & \textbf{84} & 80 & 78 & 80 \\
Towers of Hanoi (unseen) & \textbf{24} & 18 & 14 & 20 & 17 \\
\midrule
\rowcolor{wftotalrow} \textbf{Successes / 1,800} & 611 & \textbf{661} & 609 & 637 & 611 \\
\rowcolor{wfaveragerow} \textbf{Success (\%)} & 33.94 & \textbf{36.72} & 33.83 & 35.39 & 33.94 \\
\bottomrule
\end{tabularx}
\caption{Training-signal ablations on CLIPort (\%). NoPred removes future-consequence prediction supervision; Frozen blocks updates to the predictive WAM during outcome learning; NoLang removes language conditioning. The matched reference is Core, without the extra comparison head. Each setting has 100 trials.}
\label{tab:cliport-objectives}
\end{table}

%% file: tables/cliport_backbones.tex
\begin{table}[!htb]
\centering
\footnotesize
\setlength{\tabcolsep}{2.9pt}
\renewcommand{\arraystretch}{1.10}
\begin{tabularx}{\linewidth}{@{}>{\raggedright\arraybackslash}XPrrrrrQO@{}}
\toprule
\rowcolor{wfheader} \textbf{Task setting} & \textbf{Direct} & \textbf{DINO} & \textbf{SigLIP} & \textbf{\makecell{Fixed\\fusion}} & \textbf{\makecell{Learned\\fusion}} & \textbf{\makecell{Spatial\\fusion}} & \textbf{Core} & \textbf{Full} \\
\midrule
Align rope & 7 & 4 & 2 & 6 & 9 & \textbf{10} & \textbf{10} & 8 \\
Assembling kits (seen) & 20 & 10 & 20 & 23 & \textbf{25} & 24 & 24 & 22 \\
Assembling kits (unseen) & 2 & 0 & 0 & 2 & 1 & \textbf{4} & 3 & \textbf{4} \\
Packing boxes (seen) & 50 & \textbf{54} & 53 & \textbf{54} & 53 & \textbf{54} & \textbf{54} & 52 \\
Packing boxes (unseen) & 14 & 18 & 20 & \textbf{21} & 18 & \textbf{21} & 20 & 20 \\
Google objects, group (seen) & 21 & 22 & \textbf{30} & 22 & 20 & 22 & 22 & 24 \\
Google objects, seq. (seen) & 17 & \textbf{23} & 16 & 21 & 19 & 20 & 22 & 18 \\
Packing shapes & 36 & 36 & 25 & 43 & 43 & 40 & \textbf{44} & 43 \\
Google objects, group (unseen) & 20 & \textbf{25} & 19 & 22 & 22 & 18 & 20 & 22 \\
Google objects, seq. (unseen) & \textbf{7} & 5 & \textbf{7} & 2 & 6 & 3 & 3 & 5 \\
Block in bowl (seen) & \textbf{100} & 97 & \textbf{100} & \textbf{100} & \textbf{100} & \textbf{100} & \textbf{100} & \textbf{100} \\
Block in bowl (unseen) & 37 & 40 & 49 & 48 & 48 & \textbf{51} & 46 & 50 \\
Separating piles (seen) & \textbf{64} & \textbf{64} & 61 & 58 & 61 & 61 & 63 & \textbf{64} \\
Separating piles (unseen) & 41 & 37 & \textbf{50} & 42 & 44 & 45 & 42 & 44 \\
Block pyramid (seen) & 80 & 81 & \textbf{89} & 86 & 88 & 86 & 86 & 86 \\
Block pyramid (unseen) & \textbf{0} & \textbf{0} & \textbf{0} & \textbf{0} & \textbf{0} & \textbf{0} & \textbf{0} & \textbf{0} \\
Towers of Hanoi (seen) & 71 & 77 & 78 & \textbf{85} & 82 & 82 & 84 & 84 \\
Towers of Hanoi (unseen) & \textbf{24} & \textbf{24} & 22 & 16 & 18 & 18 & 18 & 17 \\
\midrule
\rowcolor{wftotalrow} \textbf{Successes} & 611 & 617 & 641 & 651 & 657 & 659 & 661 & \textbf{663} \\
\rowcolor{wfaveragerow} \textbf{Success (\%)} & 33.94 & 34.28 & 35.61 & 36.17 & 36.50 & 36.61 & 36.72 & \textbf{36.83} \\
\bottomrule
\end{tabularx}
\caption{Visual representations and fusion variants: task success (\%), 100 trials per setting. DINO and SigLIP are individual consequence models. Fixed and learned fusion combine V-JEPA2 with DINOv2; spatial fusion combines V-JEPA2 and $\pi_{0.5}$ SigLIP features. Core and Full use three V-JEPA2 WAMs.}
\label{tab:cliport-backbones}
\end{table}

%% file: sections/6.limitations.tex
\section{Limitations}\label{sec:limitations}

Candidate coverage sets the scope of improvement available to EmbodiedSkills. Its success depends on both the actions proposed by the frozen policy and the continuation they induce. Shared encoders and training data can also produce correlated prediction errors across WAM members, contributing to harmful replacements. Broader proposal coverage and more reliable consequence estimates are complementary directions for improving selection.

Performance varies across tasks. A task-cluster bootstrap with 100,000 draws quantifies sensitivity to benchmark composition: the 95\% intervals for the success-rate gains are $[-1.32,6.16]$ percentage points on MT50 and $[0.78,5.11]$ on CLIPort. The MT50 interval spans zero, reflecting the uneven task-level gains in Table~\ref{tab:metaworld-tasks}.

The current evaluations use simulated manipulation. MT50 covers repeated chunk selection, while CLIPort and CALVIN evaluate a single intervention followed by the frozen policy. Extending repeated selection to longer task sequences remains an open direction. Training also relies on restoring simulation states to collect matched candidate branches; obtaining comparable physical-robot data requires additional effort to reset objects and robot state. Real-world extensions will need to address this collection cost, visual and contact-dynamics shifts, and execution safety.

\FloatBarrier

%% file: sections/7.conclusion.tex
\section{Conclusion}\label{sec:conclusion}

We presented EmbodiedSkills, a skill-based framework that augments frozen vision--language--action policies with world action models. Five stages organize scene observation, context encoding, candidate generation, consequence prediction, outcome evaluation, candidate comparison, and action execution. These skills exchange candidate-indexed observations, predictions, and decisions through a shared feedback loop. Independent predictors compare the latent effects of the same candidate actions, and a lightweight consequence-comparison head refines selection relative to the policy's direct proposal. This organization accommodates continuous action chunks and spatial manipulation primitives while preserving the correspondence between a proposed action, its predicted effect, and the action executed in the environment.

Experiments on MetaWorld MT50 and CLIPort increase complete-task success from 76.28\% to 78.64\% and from 33.94\% to 36.83\%, respectively, with the base policies frozen. Additional comparisons place the framework alongside geometric selection, learned evaluators, and execution adapters, while CALVIN extends it to FLOWER and increases five-task-chain success from 77.10\% to 77.90\%. The ablations identify future-consequence supervision, language conditioning, joint predictive learning, and ensemble disagreement as important components of the measured improvement. Alternative representation systems fit the same workflow, while task-level results expose both successful corrections and harmful interventions. Together, these results support predictive action selection as a practical way to extend existing robot policies through reusable skills without retraining their action-generation backbone.

%% file: sections/8.detailed_results.tex
\section{Complete Task-Level Ablations}\label{sec:full-ablations}

This appendix reports the full task breakdown of the ensemble, representation, and optimization studies. Every CLIPort setting has 100 evaluation instances, so its success count equals its percentage success rate. The default Core column is shared across the studies; Full adds the consequence-comparison head and is reported separately in the main results. Bold values mark the best result within each displayed task row, including ties. Comparisons use complete-task success.

\input{tables/cliport_ensembles}
\input{tables/cliport_representation_sensitivity}
\input{tables/cliport_optimization_sensitivity}
\input{tables/cliport_all_outcomes}

\FloatBarrier

%% file: tables/cliport_ensembles.tex
\begin{table}[!htb]
\centering
\footnotesize
\setlength{\tabcolsep}{2.9pt}
\renewcommand{\arraystretch}{1.10}
\begin{tabularx}{\linewidth}{@{}>{\raggedright\arraybackslash}XPrrrrrrrO@{}}
\toprule
\rowcolor{wfheader} \textbf{Task setting} & \textbf{Direct} & \textbf{$M_1$} & \textbf{$M_2$} & \textbf{$M_3$} & \textbf{1+2} & \textbf{1+3} & \textbf{2+3} & \textbf{Mean} & \textbf{Core} \\
\midrule
Align rope & 7 & 5 & 6 & 7 & 7 & 5 & 8 & 4 & \textbf{10} \\
Assembling kits (seen) & 20 & 16 & 22 & 22 & 22 & 22 & 20 & 18 & \textbf{24} \\
Assembling kits (unseen) & 2 & 0 & 2 & 1 & 0 & 2 & \textbf{3} & 0 & \textbf{3} \\
Packing boxes (seen) & 50 & 52 & 51 & 52 & 51 & 52 & 50 & 51 & \textbf{54} \\
Packing boxes (unseen) & 14 & 21 & 24 & 21 & 21 & 17 & 20 & \textbf{26} & 20 \\
Google objects, group (seen) & 21 & 28 & 23 & 20 & 27 & \textbf{30} & 18 & 24 & 22 \\
Google objects, seq. (seen) & 17 & 18 & 19 & 16 & 20 & 16 & 18 & \textbf{23} & 22 \\
Packing shapes & 36 & 32 & 39 & 38 & 40 & 33 & 42 & 36 & \textbf{44} \\
Google objects, group (unseen) & 20 & 15 & 17 & 20 & \textbf{26} & 24 & 16 & 16 & 20 \\
Google objects, seq. (unseen) & \textbf{7} & 3 & 3 & 1 & 2 & 2 & 4 & 6 & 3 \\
Block in bowl (seen) & \textbf{100} & 96 & \textbf{100} & \textbf{100} & \textbf{100} & 96 & \textbf{100} & \textbf{100} & \textbf{100} \\
Block in bowl (unseen) & 37 & 46 & \textbf{50} & 48 & 48 & 47 & 49 & \textbf{50} & 46 \\
Separating piles (seen) & \textbf{64} & 62 & 60 & 57 & 62 & 62 & 62 & 63 & 63 \\
Separating piles (unseen) & 41 & \textbf{48} & 47 & 42 & 47 & 44 & 43 & 42 & 42 \\
Block pyramid (seen) & 80 & 83 & 85 & 78 & 84 & \textbf{86} & \textbf{86} & \textbf{86} & \textbf{86} \\
Block pyramid (unseen) & \textbf{0} & \textbf{0} & \textbf{0} & \textbf{0} & \textbf{0} & \textbf{0} & \textbf{0} & \textbf{0} & \textbf{0} \\
Towers of Hanoi (seen) & 71 & \textbf{87} & 85 & 80 & 86 & 83 & 85 & 83 & 84 \\
Towers of Hanoi (unseen) & \textbf{24} & 17 & 14 & 17 & 14 & 16 & 21 & 15 & 18 \\
\midrule
\rowcolor{wftotalrow} \textbf{Successes} & 611 & 629 & 647 & 620 & 657 & 637 & 645 & 643 & \textbf{661} \\
\rowcolor{wfaveragerow} \textbf{Success (\%)} & 33.94 & 34.94 & 35.94 & 34.44 & 36.50 & 35.39 & 35.83 & 35.72 & \textbf{36.72} \\
\bottomrule
\end{tabularx}
\caption{Complete ensemble ablation by task (\%). $M_1,M_2,M_3$ denote the three independently initialized WAMs. All three two-member pairs use mean-minus-standard-deviation scoring. Mean is the arithmetic mean of all three members; Core subtracts their population standard deviation. All members and pairs are reported.}
\label{tab:cliport-ensembles}
\end{table}

%% file: tables/cliport_representation_sensitivity.tex
\begin{table}[!htb]
\centering
\small
\setlength{\tabcolsep}{5pt}
\renewcommand{\arraystretch}{1.10}
\begin{tabularx}{\linewidth}{@{}>{\raggedright\arraybackslash}XPOrrrr@{}}
\toprule
\rowcolor{wfheader} \textbf{Task setting} & \textbf{Direct} & \textbf{Core} & \textbf{PCA} & \textbf{\makecell{Whitened\\PCA}} & \textbf{$H{=}64$} & \textbf{$H{=}256$} \\
\midrule
Align rope & 7 & \textbf{10} & 8 & 6 & 5 & 9 \\
Assembling kits (seen) & 20 & \textbf{24} & 20 & 20 & 18 & 17 \\
Assembling kits (unseen) & 2 & 3 & 2 & 3 & \textbf{4} & 0 \\
Packing boxes (seen) & 50 & \textbf{54} & \textbf{54} & 48 & \textbf{54} & 52 \\
Packing boxes (unseen) & 14 & 20 & 19 & \textbf{22} & 20 & 18 \\
Google objects, group (seen) & 21 & 22 & 22 & 24 & 19 & \textbf{27} \\
Google objects, seq. (seen) & 17 & \textbf{22} & 16 & 16 & 20 & \textbf{22} \\
Packing shapes & 36 & \textbf{44} & 36 & 38 & 39 & 40 \\
Google objects, group (unseen) & 20 & 20 & 16 & \textbf{22} & 12 & 14 \\
Google objects, seq. (unseen) & \textbf{7} & 3 & 5 & 4 & 0 & 0 \\
Block in bowl (seen) & \textbf{100} & \textbf{100} & 97 & 95 & 96 & \textbf{100} \\
Block in bowl (unseen) & 37 & 46 & 46 & 44 & \textbf{48} & 44 \\
Separating piles (seen) & 64 & 63 & 61 & 61 & \textbf{66} & 60 \\
Separating piles (unseen) & 41 & 42 & 39 & 39 & 43 & \textbf{46} \\
Block pyramid (seen) & 80 & \textbf{86} & \textbf{86} & 83 & 81 & 78 \\
Block pyramid (unseen) & \textbf{0} & \textbf{0} & \textbf{0} & \textbf{0} & \textbf{0} & \textbf{0} \\
Towers of Hanoi (seen) & 71 & \textbf{84} & 80 & 76 & \textbf{84} & 78 \\
Towers of Hanoi (unseen) & \textbf{24} & 18 & 14 & 8 & 20 & 18 \\
\midrule
\rowcolor{wftotalrow} \textbf{Successes} & 611 & \textbf{661} & 621 & 609 & 629 & 623 \\
\rowcolor{wfaveragerow} \textbf{Success (\%)} & 33.94 & \textbf{36.72} & 34.50 & 33.83 & 34.94 & 34.61 \\
\bottomrule
\end{tabularx}
\caption{Projection and predictor-capacity sensitivity (\%). Core uses a fixed random orthogonal $1024\!\rightarrow\!32$ projection and hidden width $H=128$. PCA and whitened PCA are fitted only on training transitions. Each alternative changes one factor relative to Core; 100 trials per task.}
\label{tab:cliport-representation}
\end{table}

%% file: tables/cliport_optimization_sensitivity.tex
\begin{table}[!htb]
\centering
\small
\setlength{\tabcolsep}{5pt}
\renewcommand{\arraystretch}{1.10}
\begin{tabularx}{\linewidth}{@{}>{\raggedright\arraybackslash}XPOrrrr@{}}
\toprule
\rowcolor{wfheader} \textbf{Task setting} & \textbf{Direct} & \textbf{Core} & \textbf{\makecell{LR\\$10^{-4}$}} & \textbf{\makecell{LR\\$4\!\times\!10^{-4}$}} & \textbf{$\lambda{=}0.5$} & \textbf{$\lambda{=}2$} \\
\midrule
Align rope & 7 & \textbf{10} & 4 & 4 & 9 & 6 \\
Assembling kits (seen) & 20 & \textbf{24} & 22 & 15 & 20 & 19 \\
Assembling kits (unseen) & 2 & \textbf{3} & 0 & 0 & 2 & 0 \\
Packing boxes (seen) & 50 & 54 & \textbf{55} & 50 & 48 & 54 \\
Packing boxes (unseen) & 14 & 20 & \textbf{23} & 18 & \textbf{23} & 22 \\
Google objects, group (seen) & 21 & 22 & \textbf{26} & 22 & 24 & 25 \\
Google objects, seq. (seen) & 17 & \textbf{22} & 20 & 14 & \textbf{22} & 18 \\
Packing shapes & 36 & \textbf{44} & 34 & 32 & 34 & 40 \\
Google objects, group (unseen) & 20 & 20 & 22 & 21 & \textbf{23} & 16 \\
Google objects, seq. (unseen) & \textbf{7} & 3 & 3 & 6 & 4 & 1 \\
Block in bowl (seen) & \textbf{100} & \textbf{100} & 99 & \textbf{100} & \textbf{100} & \textbf{100} \\
Block in bowl (unseen) & 37 & 46 & 46 & \textbf{48} & 47 & 45 \\
Separating piles (seen) & 64 & 63 & 65 & \textbf{66} & 65 & 60 \\
Separating piles (unseen) & 41 & \textbf{42} & 40 & 39 & 40 & 40 \\
Block pyramid (seen) & 80 & \textbf{86} & 80 & 80 & 84 & 83 \\
Block pyramid (unseen) & \textbf{0} & \textbf{0} & \textbf{0} & \textbf{0} & \textbf{0} & \textbf{0} \\
Towers of Hanoi (seen) & 71 & \textbf{84} & 80 & \textbf{84} & 80 & 80 \\
Towers of Hanoi (unseen) & \textbf{24} & 18 & 14 & 18 & 16 & 22 \\
\midrule
\rowcolor{wftotalrow} \textbf{Successes} & 611 & \textbf{661} & 633 & 617 & 641 & 631 \\
\rowcolor{wfaveragerow} \textbf{Success (\%)} & 33.94 & \textbf{36.72} & 35.17 & 34.28 & 35.61 & 35.06 \\
\bottomrule
\end{tabularx}
\caption{Optimization sensitivity (\%). Core uses a joint-stage WAM learning rate of $2\times10^{-4}$ and a consequence-loss weight $\lambda=1$. The value-head learning rate stays fixed at $2\times10^{-4}$. All entries use 100 trials per task.}
\label{tab:cliport-optimization}
\end{table}

%% file: tables/cliport_all_outcomes.tex
\begin{table}[!htb]
\centering
\small
\setlength{\tabcolsep}{6pt}
\renewcommand{\arraystretch}{1.10}
\begin{tabularx}{\linewidth}{@{}>{\raggedright\arraybackslash}XrOArrr@{}}
\toprule
\rowcolor{wfheader} \textbf{Configuration} & \textbf{Successes} & \textbf{Rate} & \textbf{$\Delta N$} & \textbf{Rescue} & \textbf{Harm} & \textbf{Switch} \\
\midrule
Direct & 611 & 33.94\% & \same{0} & 0 & 0 & 0 \\
\rowcolor{wftotalrow} Core & 661 & 36.72\% & \gain{+50} & 87 & 37 & 986 \\
\rowcolor{wfaveragerow} Full & 663 & 36.83\% & \gain{+52} & 92 & 40 & 978 \\
$K=2$ & 639 & 35.50\% & \gain{+28} & 40 & 12 & 497 \\
$K=3$ & 645 & 35.83\% & \gain{+34} & 54 & 20 & 714 \\
Single $M_1$ & 629 & 34.94\% & \gain{+18} & 87 & 69 & 1493 \\
Single $M_2$ & 647 & 35.94\% & \gain{+36} & 104 & 68 & 1477 \\
Single $M_3$ & 620 & 34.44\% & \gain{+9} & 90 & 81 & 1500 \\
Pair $M_1+M_2$ & 657 & 36.50\% & \gain{+46} & 98 & 52 & 1153 \\
Pair $M_1+M_3$ & 637 & 35.39\% & \gain{+26} & 76 & 50 & 1192 \\
Pair $M_2+M_3$ & 645 & 35.83\% & \gain{+34} & 87 & 53 & 1176 \\
Three-member mean & 643 & 35.72\% & \gain{+32} & 103 & 71 & 1506 \\
NoPred & 609 & 33.83\% & \loss{-2} & 57 & 59 & 919 \\
Frozen predictor & 637 & 35.39\% & \gain{+26} & 58 & 32 & 795 \\
NoLang & 611 & 33.94\% & \same{0} & 59 & 59 & 1087 \\
DINOv2-L & 617 & 34.28\% & \gain{+6} & 74 & 68 & 1412 \\
SigLIP & 641 & 35.61\% & \gain{+30} & 91 & 61 & 1511 \\
Fixed V-JEPA2/DINO fusion & 651 & 36.17\% & \gain{+40} & 83 & 43 & 1004 \\
Learned V-JEPA2/DINO fusion & 657 & 36.50\% & \gain{+46} & 87 & 41 & 1004 \\
Spatial V-JEPA2/SigLIP fusion & 659 & 36.61\% & \gain{+48} & 88 & 40 & 1004 \\
PCA32 & 621 & 34.50\% & \gain{+10} & 64 & 54 & 954 \\
Whitened PCA32 & 609 & 33.83\% & \loss{-2} & 63 & 65 & 1108 \\
Hidden 64 & 629 & 34.94\% & \gain{+18} & 65 & 47 & 954 \\
Hidden 256 & 623 & 34.61\% & \gain{+12} & 57 & 45 & 883 \\
Learning rate $10^{-4}$ & 633 & 35.17\% & \gain{+22} & 68 & 46 & 948 \\
Learning rate $4\times10^{-4}$ & 617 & 34.28\% & \gain{+6} & 56 & 50 & 1053 \\
Consequence weight 0.5 & 641 & 35.61\% & \gain{+30} & 76 & 46 & 985 \\
Consequence weight 2.0 & 631 & 35.06\% & \gain{+20} & 63 & 43 & 938 \\
\bottomrule
\end{tabularx}
\caption{Aggregate CLIPort outcomes for every distinct reported configuration (1,800 trials). Rescue changes a Direct failure into success; Harm changes a Direct success into failure. Switch counts all replacements of candidate 0, including those with unchanged terminal outcomes. Thus $\Delta N=\mathrm{Rescue}-\mathrm{Harm}$.}
\label{tab:cliport-all-outcomes}
\end{table}

%% file: sections/9.expanded_comparisons.tex
\section{Additional Policy Comparisons}\label{sec:expanded-comparison-protocols}

\subsection{Method Adaptations and Evaluation Scope}

The comparisons in Table~\ref{tab:policy-augmentation} cover 2,500 MT50 episodes (50 tasks with 50 episodes each) and 1,800 CLIPort instances (18 settings with 100 instances each). These are the evaluation sizes used in Table~\ref{tab:wam-overall}. The following adaptations retain the author algorithms or identified components while accommodating the local action and observation interfaces. Complete Direct and Full task breakdowns appear in Tables~\ref{tab:metaworld-tasks} and~\ref{tab:cliport-tasks}.

\paragraph{Geometric and learned candidate selection.}
KeyStone~\cite{dai2026geometryguidedselfconsistencyphysical} uses the author guarded cluster-medoid selector with five candidates, two clusters, and the released threshold 0.3. On CLIPort, each candidate is represented by its two pick/place pose endpoints, including position and quaternion coordinates. Distances use the author Euclidean metric over these spatial hypotheses. TACO~\cite{yang2025taco} uses the author coin-flipping network component on the MT50 policy features. The GAF comparison~\cite{yang2026guidedactionflowvalueguided} uses the fixed earlier MLP-ensemble implementation.

\paragraph{Verifier and value components.}
UF-OPS~\cite{attarian2026updatefreeonpolicysteeringverifiers} reuses the author contrastive classifier, time embedding, and binary-plus-contrastive training objective. Its observation and action input widths are adapted to the frozen V-JEPA~2 and MiniLM features and the native CLIPort or FLOWER actions. This verifier component selects among the five supplied candidates. SPIRAL, the policy-improvement method introduced with SARM2~\cite{chen2026sarm2multitaskstageaware}, supplies the author five-critic ensemble, minimum-value aggregation, and sparse Monte Carlo objective. Supervision is discounted terminal success under the recorded continuation. We evaluate this critic as a standalone selector over frozen-policy proposals. Both components train on the corresponding branch-training partitions, with checkpoint selection on validation data. Deployed scoring uses current observation features and proposed actions.

\paragraph{Action correction and execution.}
VLA-Corrector~\cite{pan2026vlacorrector} retains the author latent monitor and triggered guidance, adapted to eight-step visual transitions and the OpenPI interface. ASRR-C~\cite{asrr2026} uses the author context-conditioned residual refiner with hidden width 256, trained against expert action suffixes. RTC~\cite{black2025realtimeexecutionactionchunking} retains the real-time resampling rule with a four-action execution block and two-step simulated delay. VLA-RAIL~\cite{zhao2025vlarailrealtimeasynchronousinference} uses its smoothing and inter-chunk fusion modules in a synchronous Cartesian-control adapter. AutoHorizon~\cite{wang2026vlaknowslimitsadaptive} retains attention-based horizon estimation, allowing one to ten executed actions. A$^3$~\cite{chen2026dynamicexecutioncommitmentvisionlanguageaction} uses author speculative acceptance with eight drafts and a five-step speculative horizon, dispatching at most eight accepted controls. The latter two methods use the public policy through their required attention-enabled or converted implementations. These MT50 system comparisons retain the respective action-generation and execution schedules.

\subsection{CALVIN Chain Protocol}\label{sec:calvin-chain-protocol}

The CALVIN adaptation retains the three consequence predictors, four-outcome comparison head, and fixed selection rule. Its training archive contains 194 usable anchors and its validation archive 69, each spanning all 34 task labels. The FLOWER backend produces five ten-step chunks in its native seven-dimensional action space. Inputs are frozen visual and language features, the observed 15-dimensional robot state, and elapsed control progress.

The evaluation contains 1,000 five-task chains. Each is assigned one intervention task and a predetermined control point. The direct policy executes the prefix; all candidates then start from the matched intervention state and are followed by the frozen policy to chain termination. A candidate's training outcome describes success of the entire chain, while its prediction target describes the immediate ten-step transition. The Direct comparator uses candidate~0 with the same policy reset at the branch boundary.

All scheduled chains remain in the denominator, including failures before the assigned intervention. SR$k$ divides the number of chains completing at least $k$ tasks by 1,000. Full's completion counts for $k=1,\ldots,5$ are 995, 963, 917, 868, and 779; Direct's are 995, 962, 917, 864, and 771. Summing these counts and dividing by 1,000 gives average completed lengths of 4.522 and 4.509, respectively. The current task instruction conditions candidate evaluation; terminal branch outcomes supply the supervision and evaluation targets.

\FloatBarrier